# Time-Series Classification in Smart Manufacturing Systems: An Experimental Evaluation of State-of-the-Art Machine Learning Algorithms

Mojtaba A. Farahani[a], M. R. McCormick[a], Ramy Harik[b], and Thorsten Wuest[a1]

[a] West Virginia University, Morgantown, 26505 WV, U.S.A

[b] University of South Carolina, Columbia, 29208, SC, U.S.A

# Abstract

Manufacturing is transformed towards smart manufacturing, entering a new data-driven era fueled by digital technologies. The resulting Smart Manufacturing Systems (SMS) gather extensive amounts of diverse data, thanks to the growing number of sensors and rapid advances in sensing technologies. Among the various data types available in SMS settings, time-series data plays a pivotal role. Hence, Time-Series Classification (TSC) emerges as a crucial task in this domain. Over the past decade, researchers have introduced numerous methods for TSC, necessitating not only algorithmic development and analysis but also validation and empirical comparison. This dual approach holds substantial value for practitioners by streamlining choices and revealing insights into models' strengths and weaknesses. The objective of this study is to fill this gap by providing a rigorous experimental evaluation of the state-of-the-art Machine Learning (ML) and Deep Learning (DL) algorithms for TSC tasks in manufacturing and industrial settings. We first explored and compiled a comprehensive list of more than 92 state-of-the-art algorithms from both TSC and manufacturing literature. Following this, we methodologically selected the 36 most representative algorithms from this list. To evaluate their performance across various manufacturing classification tasks, we curated a set of 22 manufacturing datasets, representative of different characteristics that cover diverse manufacturing problems. Subsequently, we implemented and evaluated the algorithms on the manufacturing benchmark datasets, and analyzed the results for each dataset. Based on the results, ResNet, DrCIF, InceptionTime, and ARSENAL emerged as the top-performing algorithms, boasting an average accuracy of over 96.6% across all 22 manufacturing TSC datasets. These findings underscore the robustness, efficiency, scalability, and effectiveness of convolutional kernels in capturing temporal features in time-series data collected from manufacturing systems for TSC tasks, as three out of the top four performing algorithms leverage these kernels for feature extraction. Additionally, LSTM, BiLSTM, and TS-LSTM algorithms deserve recognition for their effectiveness in capturing features within manufacturing time-series data using RNN-based structures.

**Keywords:** Smart Manufacturing, Industry 4.0, Time-Series Classification, Machine Learning, AI

[1] Corresponding Author: thwuest@mail.wvu.edu (T. Wuest)

# 1. Introduction

Manufacturing is undergoing a transformation towards a new digital and data-driven era known as Industry 4.0 and smart manufacturing systems. This digital transformation is defined by the core principles of *connectivity*, *virtualization*, and *data utilization*[1]. Smart manufacturing technologies enable manufacturers to collect large amounts of data and provide the tools to derive insights from these often massive and diverse datasets using Machine Learning (ML) and Artificial Intelligence (AI). *Data utilization* specifically entails translating and contextualizing collected data into actionable insights through advanced analytics. This assists in the data-driven revolution proposed by smart manufacturing. Specific tools to power this revolution include AI, ML, and Digital Twins (DT) that leverage the increasingly available amount of data provided by manufacturing systems and processes. As the needs of manufacturers increase in an ever-competitive landscape, the use of AI and ML approaches, techniques, and algorithms accelerates[2,3].

ML has been applied in a wide range of manufacturing and industrial settings to address a variety of diverse problems by providing enhanced analytics capabilities. Examples of ML applications in manufacturing include but are not limited to image-based quality inspection[4], fault detection and diagnosis[5,6], process optimization[7], energy cost reduction[8], customer demand forecasting, and real-time machine monitoring for Remaining Useful Life (RUL) forecasting[9]. Time-series data is one of the main data types that are available in manufacturing settings[10]. It is becoming more ubiquitous thanks to the increasing number of sensors and sensing technologies and can be found across a wide range of industries outside of manufacturing. Stock market prices in financial markets[11], ECG data in healthcare[12], Ecohydrology sensing data[13], positional data from smart wearable devices, high-resolution images from the sun over time[14], 3D depth sensor Kinect data[15], and vibration, pressure, and temperature data coming from manufacturing sensors[16] are all examples of time-series data.

The general goal of time-series analytics is to approximate a dataset in terms of understanding the underlying relation between data points in the time-series and incorporating consideration of recognized patterns into generated predictions. Time-series analytics is considered one of the most challenging problems in data mining mainly because of temporal dependencies, potential variable lengths, potential seasonality, trend, non-stationarity, and noise. In general, having ordered values adds a layer of complexity to a problem[17–19]. The primary types of time-series analysis are *time-series classification*, *time-series forecasting, anomaly detection*, and *clustering*[17]. *Time-Series Classification (TSC)* is a predictive task that leverages supervised learning approaches to learn from labeled data and categorize them into labeled classes. *Time-series forecasting* seeks to understand data components (such as trends, seasonality, and cycles) to predict future behaviors and values. Finally, *time-series clustering* and *anomaly detection* tasks, often grouped together, use unsupervised learning approaches to create groups or clusters of data with similar properties and/or to detect anomalous data.

Manufacturing presents a unique opportunity for leveraging time-series analytics with the increasing prevalence of industrial sensors. Examples of TSC applications in smart manufacturing systems include but are not limited to quality inspection and control, predictive maintenance, supply chain optimization, and energy management. With the growth of Industry 4.0 and smart manufacturing, new machine tools are already equipped with advanced sensing technologies, and existing legacy systems are rapidly outfitted with large amounts of new and powerful sensors. These sensors are capable of automatically accumulating various time-series data[20]. These sensors connect the physical assets on the shop floor and beyond to a digital network using the Industrial Internet of Things (IIoT) to collect and share data. Combining this increasing quantity and quality of data with abilities to derive meaningful insights provides organizations with the ability to develop a sustained competitive advantage. This business case advances the need for researchers to investigate methods to maximize the impact of time-series analytics and at the same time, the need to ensure the rapid transfer of new knowledge to practitioners in industry.

*The motivation* behind this work is the fact that besides focusing on algorithm development and analysis, it is essential to concurrently undertake the validation and empirical comparison of the numerous existing algorithms. This endeavor holds immense value for practitioners as it narrows their options and provides insights into the strengths and weaknesses of available models. At the same time, the manufacturing community, especially industry, is in desperate need of practical guidance on the issue. This lack of practical guidance is the main motivation for this study. To date, such an investigation has not been conducted for TSC on manufacturing data sets, and the objective of our paper is to fill this pressing research gap.

The *main focus* of this paper, as depicted in Figure 1, is to examine TSC algorithms that can be effectively and efficiently applied in the manufacturing domain. This study reviews, categorizes, and evaluates the state-of-the-art TSC algorithms on a diverse set of manufacturing problems represented by different data sets. To achieve this, first, the state-of-the-art TSC algorithms are identified and extracted from the literature before we use our novel and transparent methodology to select the most representative (state-of-the-art and baseline) TSC algorithms from the different categories. During the study, we identified a gap between the TSC algorithms that are common in Computer Science (CS) and the ones dominant in the manufacturing literature. To ensure we cover the best and most advanced of both worlds, the initial list of TSC algorithms comprises both the TSC community in CS and the manufacturing literature to ensure completeness. To identify the state-of-the-art TSC algorithms in smart manufacturing settings, we implement the down-selected algorithms on publicly available manufacturing datasets, perform an empirical comparative study on them, and carefully evaluate the results. It should be noted that in Figure 1, the size of the bubbles does not have specific meanings, and the depiction is only for illustrative purposes.

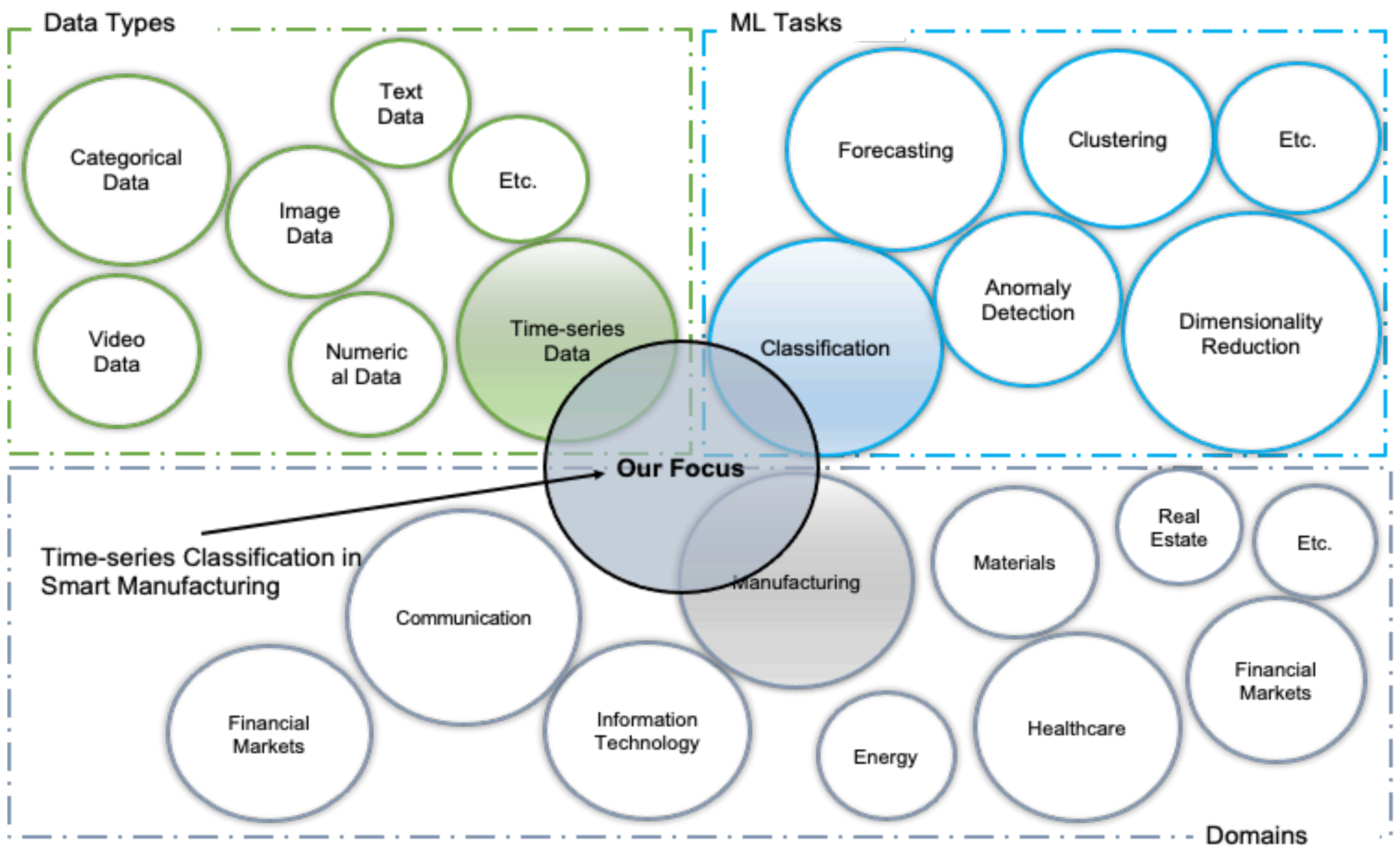

Figure 1: Graphical representation of the scope of work: Time-series classification in manufacturing

The *scope of this study* is i) to provide a list of preprocessed manufacturing-related datasets for TSC tasks, their characteristics, problem types, and a brief description of each dataset, ii) to provide a list of all TSC algorithms that have been recently proposed in both the CS and manufacturing literature, iii) curate a representative list of state-of-the-art and legacy discriminative TSC algorithms, and iv) implement an experimental evaluation of the algorithms on manufacturing datasets to identify the best-performing algorithms for each problem. It should be noted that the detailed theoretical explanation of specific algorithms is out of the scope of this study and we suggest consulting the references provided for a deeper dive if desired. To the best of our knowledge, this has not been done for the manufacturing domain. *This study's novelty* lies in starting the research in the smart manufacturing domain towards more promising avenues and providing out-of-the-box solutions for practitioners in the field. Furthermore, this work offers guidance on the application of TSC algorithms and establishes a connection between otherwise potentially isolated domains— manufacturing and general TSC in CS.

More specifically, we aim to answer the following *research questions* in this study:

- RQ1: What are the characteristics of public datasets in manufacturing and industrial settings that can be utilized for TSC tasks, and what standardized preprocessing methods can be applied to prepare them for TSC algorithms?
- RQ2: What are the characteristics of state-of-the-art algorithms for TSC tasks applicable to manufacturing datasets, and what features make these algorithms particularly well-suited for solving classification problems in the manufacturing domain?
- RQ3: Which algorithms demonstrate superior performance on the mentioned datasets and across various TSC challenges within the manufacturing domain? How can these findings inform the algorithm selection process for other users and a wide range of diverse problems?

The remainder of this paper is organized as follows: *Section Two* provides the necessary background and categorization that has been used in this research for TSC algorithms as well as publicly available datasets. In *Section Three*, we describe the experimental evaluation methodology of this study that led to the final selection of datasets and TSC algorithms, as well as the comparison and evaluation metrics. In Section *Four*, we present the results of our experiments with different scenarios based on our defined evaluation metrics as well as additional discussion, highlighted challenges, and reflection on limitations that we faced during this study. Finally, we present conclusions drawn from the previous sections and recommendations for future research in *Section Five*.

# 2. Background

In this section, we start by providing the necessary background information and definitions that will be used for the remainder of the paper. Furthermore, we will present our taxonomy to categorize TSC algorithms into several groups.

## 2.1. TSC Algorithms

It is worthwhile to reflect on the definitions for TSC that we are following throughout the paper before introducing the different types of TSC algorithms. Time-series is defined as an ordered set of real-valued numbers. A *univariate time series,* $\mathbf{x}=[x_1, x_2,..., x_T]$ is an ordered set of real values with length *T*. In contrast, a M-dimensional *multivariate time-series*, $\mathbf{X} = [x^1, x^2,..., x^M]$ is a matrix consisting of *M* different univariate time-series with $x^i \in R^i$. A *dataset* D = ($\mathbf{X}$, $\mathbf{Y}$) = {($\mathbf{X}_1$,$y_1$),($\mathbf{X}_2$,$y_2$),...,($\mathbf{X}_N$,$y_N$)} with a dimension of *(N, T, M)* is a collection of pairs ($\mathbf{X}_i$,$y_i$) where $\mathbf{X}_i$ could either be a univariate (i.e., M = 1) or multivariate (i.e., $M > 1$) time-series with $y_i$ as its corresponding supervisory label[19]. A visual representation of time-series dimensions is shown in Figure 2.

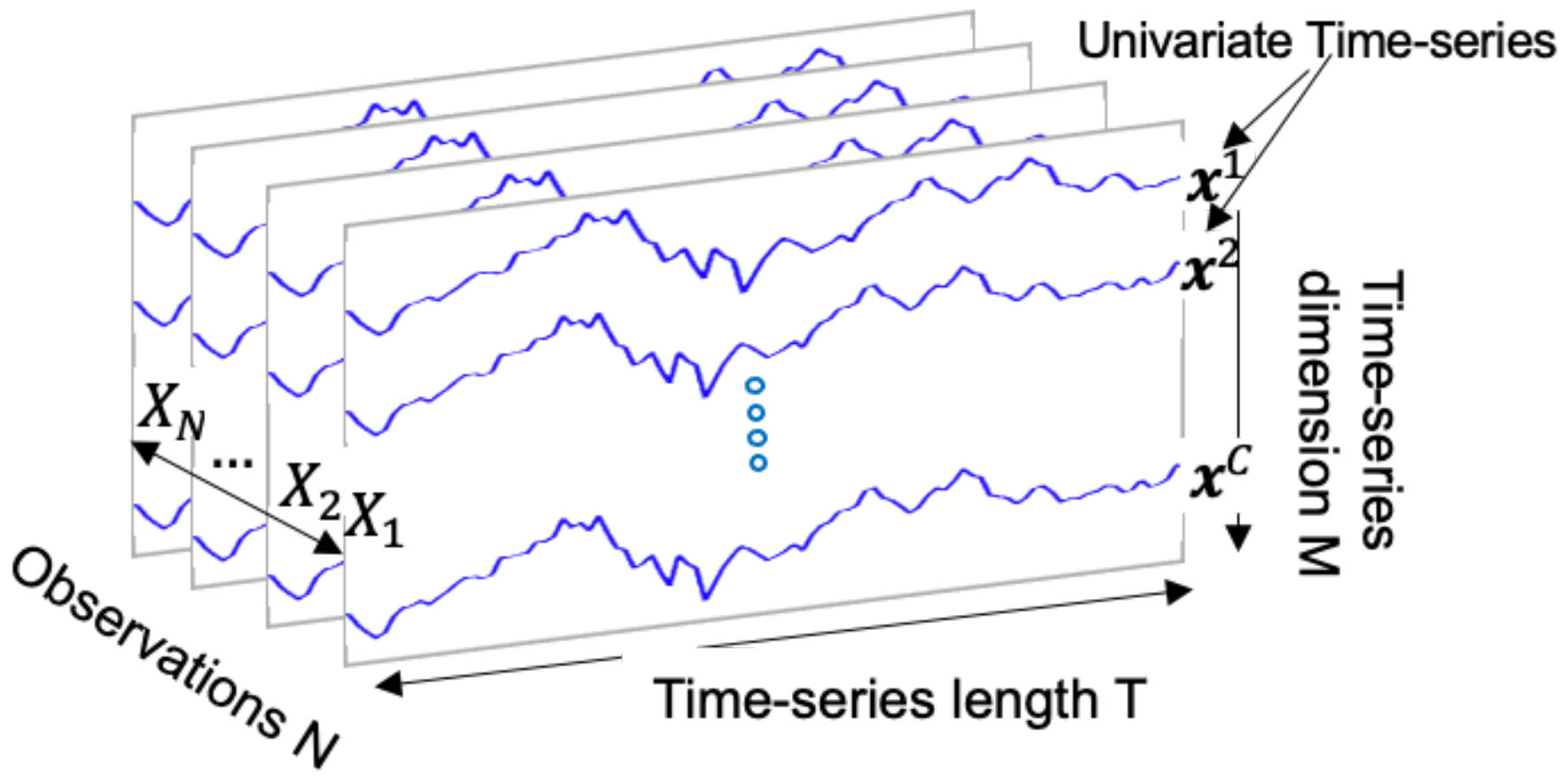

Figure 2: Visual representation of a multivariate time-series dataset with *(N, T, M)* dimension[10]

The *main task of TSC* is to train a classification algorithm on dataset *D* to find an approximation function that maps the inputs **X** to a probability distribution over the class labels. Particularly, TSC can be categorized into Univariate Time-Series Classification (UTSC) and Multivariate Time-Series Classification (MTSC). The data collected from a single sensor are referred to as univariate time-series (i.e., $M = 1$), while the data collected from multiple sensors simultaneously are referred to as multivariate time series ($M > 1$). Historically, the main focus of the TSC community has been on developing UTSC algorithms. However, in today's reality, it is more common to encounter MTSC problems[17]. Thus, there has been increasing attention on the development of MTSC algorithms for different applications. It must be noted that algorithms designed for MTSC tasks can be used for UTSC tasks, while vice versa is not necessarily the case. Traditional classification algorithms may miss important characteristics of the data due to the possibility of discriminatory features dependent on the ordering. Thus, during the last decade, researchers have proposed hundreds of methods and algorithms to overcome this problem.[21]

The general framework applied to a TSC problem is illustrated in Figure 3. It involves a series of preprocessing modules applied to the raw sensor data. These preprocessing steps yield a dataset with dimensions *(N, T, M)*, which serves as the input to the TSC module. The TSC module itself consists of two main parts: a Feature Engineering (FE) algorithm and a classifier algorithm. It's worth noting that the FE algorithm is optional, and TSC can be performed without it. The ultimate output of the TSC module is the prediction of class labels, providing valuable insights into the problem being addressed. This paper's primary focus centers on the TSC module. The TSC module is also called a TSC algorithm in the literature.

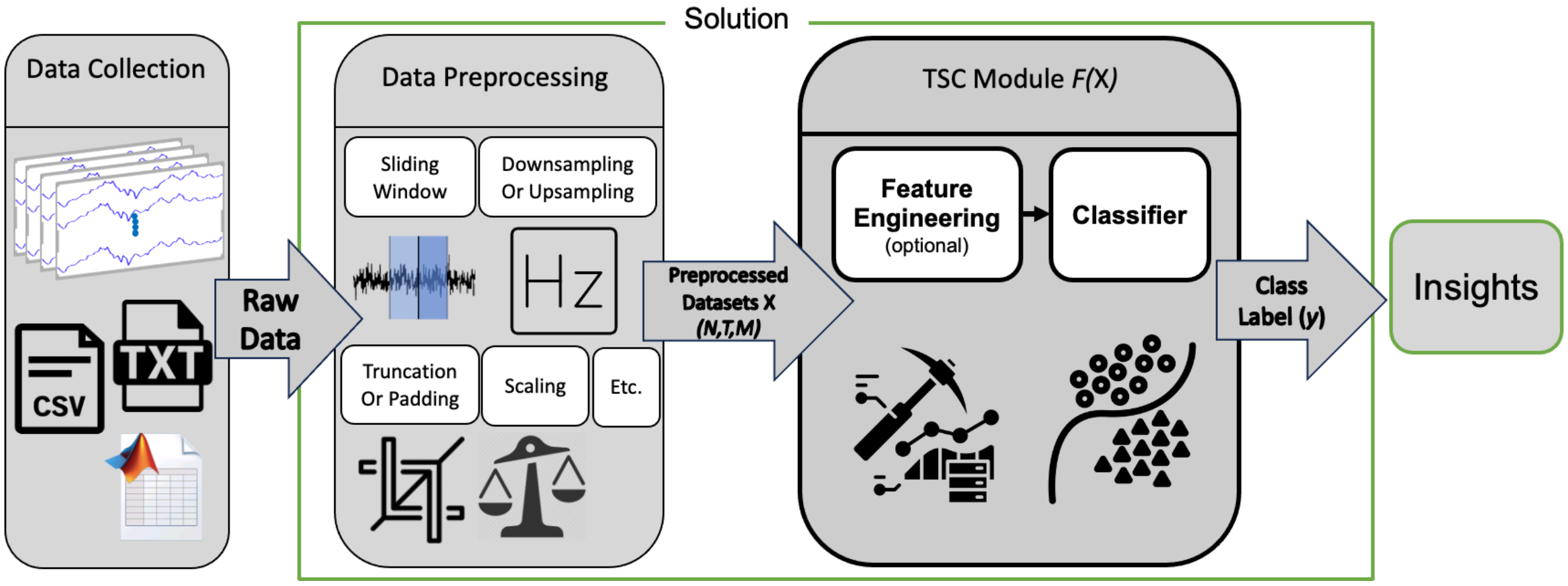

Figure 3: General framework for a TSC solution

There are two main approaches to constructing a TSC module. Conventionally, this has been addressed by defining a distance metric between time-series instances and comparing the sequential values directly. This approach is referred to as *instance-based classification* in the literature. An alternative approach is to utilize *feature-based classification***.** Theoretically, this approach utilizes FE techniques to transform the data from temporal to static, making the raw data more separable, and enhancing the solution quality. FE is the process of using domain knowledge to extract, transform, or select relevant features (characteristics, properties, and attributes) from raw data to grasp the essence of the data. A simple example is to represent a time-series using its statistical features such as mean, min, max, and variance, thereby transforming a time-series of any length into short vectors that encapsulate these properties.[10,22]

In this study, to address the research gaps, we compiled a list of algorithms specifically designed for TSC tasks from two main categories of resources. To begin with, we extracted *90 research papers* relevant to TSC in manufacturing and industrial settings from our previous work that could potentially be used as a reference for TSC tasks[10]. We analyzed all 90 papers from the proposed ML TSC algorithm in detail, resulting in *36 papers* from that list that provided sufficient information or explanation to allow the reproduction of their proposed ML algorithms. Furthermore, there are several studies from the TSC community in the CS field introducing and comparing TSC algorithms[19,21,23]. After removing duplicate algorithms that were used in multiple research studies, the result is a comprehensive list of *92 TSC algorithms* reflecting the current state-of-the-art in both manufacturing and CS. Moreover, we categorized the algorithms in a hierarchical structure that helps with decision-making and algorithm selection. To the best of our knowledge, there is no existing research providing a comprehensive list of applicable ML algorithms that can be used for TSC tasks in the manufacturing domain.

TSC techniques and algorithms that have been proposed in the literature can be categorized into two main categories, namely *conventional ML methods* and *Artificial Neural Networks (ANN) & deep learning-based methods*. DL is a specific subfield of ML where the learning happens in successive layers of feature representations in a neural network structure. There has been increasing attention on DL techniques in recent years across all ML applications and many researchers consider ANN/DL techniques as a separate category from conventional ML techniques. One reason behind that may be due to the ability of automatic feature extraction in DL techniques whereas, in conventional ML techniques, hand-crafted features should be constructed and calculated before executing any ML task. We define conventional ML as the group of non-ANN and DL techniques and algorithms that use hand-crafted features in their learning process and unlike ANN/DL algorithms, are incapable of learning features automatically. Some other authors may use the "Traditional ML" term for this group of algorithms, but we prefer "conventional ML" for the sake of consistency with our previous research. It should also be noted that due to the diversity of research and publications regarding ML across a multitude of sub-domains, industries, and applications, a variety of methods for classifying ML algorithms have emerged and resulted in a lack of consensus [3].

### 2.1.1. Conventional TSC Algorithms

Conventional ML algorithms are a well-established topic in the TSC community. Various models have been proposed to achieve this objective. Since there are two main modules in a TSC algorithm (see Figure 3), we categorize conventional ML algorithms from *two separate points of view* (i.e., FE technique and classification technique).

#### 2.1.1.1 Feature Engineering Techniques

Figure 4 illustrates nine different groups of algorithms, each applying different FE techniques on raw time-series. 'Raw' in this instance describes data sets that are not abstracted as feature sets. It is important to highlight that instance-based approaches do not employ an FE module, resulting in a 'None' categorization for FE. The numbers in parentheses correspond to the count of algorithms in each category from the initial compiled list of 92 algorithms. In this paper, we briefly introduce the algorithms, and if more details are desired, we refer the reader to the referenced papers in subsequent sections.

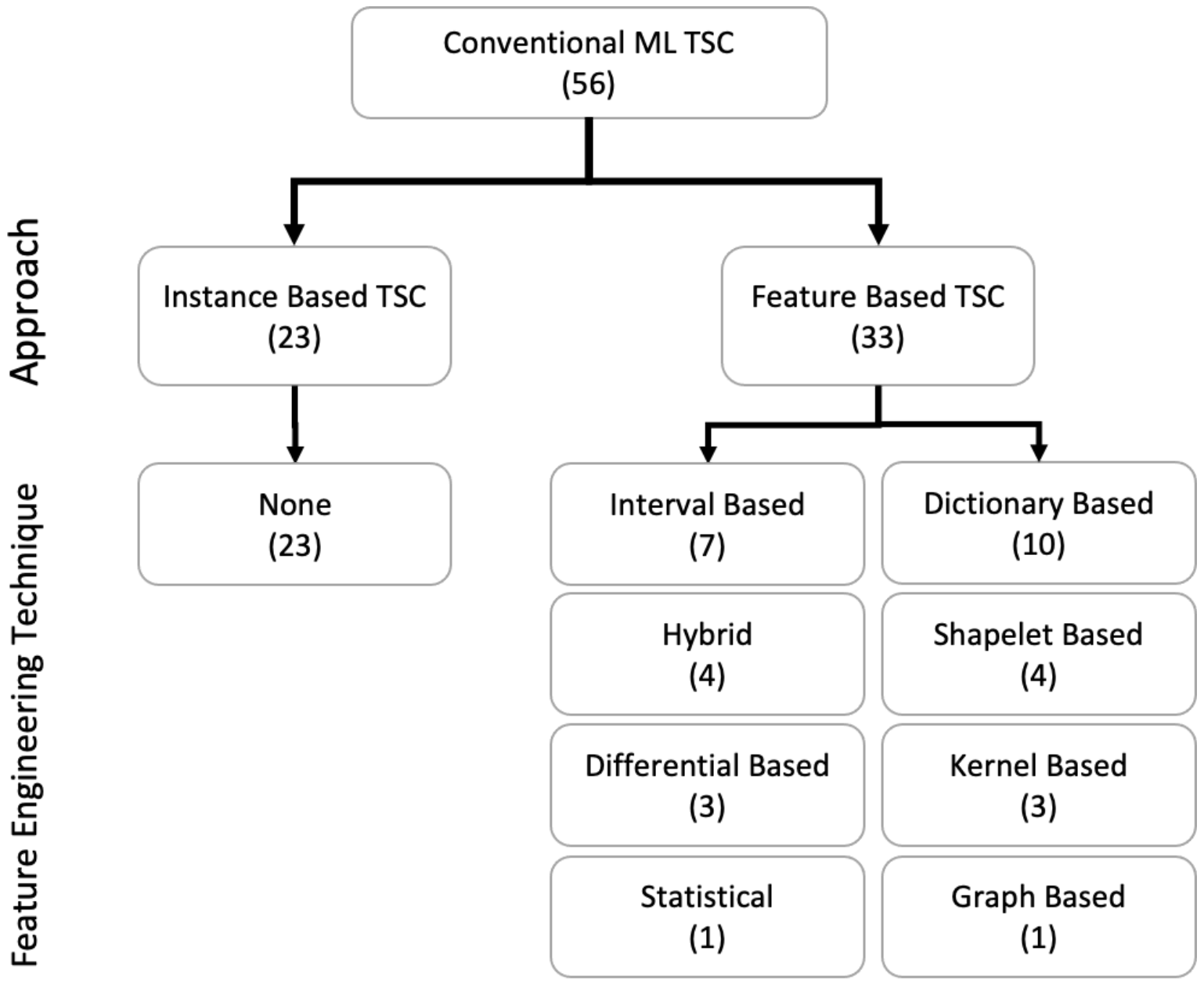

Figure 4: FE techniques in conventional ML TSC algorithms

*Dictionary-based algorithms* carry out FE by transforming the input time-series data into representative words, then basing similarity on comparing the distribution of words. Examples of dictionary-based FE algorithms include Piecewise Aggregate Approximation (PAA)[24], Symbolic Aggregate approXimation (SAX), and Symbolic Fourier Approximation (SFA)[25]. Several algorithms, such as Bag of Patterns (BOP)[21], Symbolic Aggregate Approximation Vector Space Model (SAXVSM)[26], Bag of SFA Symbols (BOSS)[27], Bag Of SFASymbols in Vector Space (BOSSVS)[28], Dynamic Time Warping Features (DTWF)[29], WEASEL+MUSE[30], Contractable Bag of Symbolic-Fourier Approximation Symbols (CBOSS)[31], the Temporal Dictionary Ensemble (TDE)[32], the multiple representation sequence learner (MrSEQL)[33], and Multiple representations SEQuence Miner (MrSQM)[33] use these dictionary-based FE techniques coupled with different kinds of classifiers (e.g., K-Nearest Neighbor (KNN), Support Vector Machines (SVM), and Logistic Regression (LR)) to perform the TSC task. Table 1 shows these algorithms, their main FE algorithms, and their respective classification techniques and algorithms.

Table 1. Conventional ML TSC algorithms using dictionary-based FE techniques.

| Algorithm Name | FE Algorithms | Classification Technique | Classification Algorithm |
|---|---|---|---|
| BOP | PAA, SAX | Distance-based | KNN |
| SAXSVM | SAX, VSM | Distance-based | KNN |
| BOSS | SFA | Distance-based | KNN |
| BOSSVS | SFA, tf-idf | Distance-based | KNN |
| DTWF | DTW, SAX | Distance-based | SVM |
| WEASEL+MUSE | SFA, first-order differences SFA | Statistical | LR |
| CBOSS | SFA | Algorithm Ensemble | BOSS |
| TDE | SFA | Algorithm Ensemble | BOSS |
| MrSEQL | SFA, SAX | Statistical | LR |
| MrSQM | SFA, SAX | Statistical | LR |

*Interval-based algorithms* are a family of algorithms that derive features from intervals of each series. For a series of length T, there are T(T−1)/2 possible contiguous intervals. These algorithms extract different kinds of features (e.g., statistical summary features, spectral features, catch22 features, HCTSA features, etc.) from time-series intervals and then train a KNN, Linear Discriminant analysis (LDA), Time-series Tree, Decision Tree (DT), or Random Forest (RF) classifier on the extracted feature to do the TSC task. Several algorithms such as Fulcher and Jones's feature-based linear classifier (FBL)[22], Time Series Bag of Features (TSBF)[21], Learned Pattern Similarity (LPS)[21], Time Series Forest (TSF)[34], Random Interval Spectral Ensemble (RISE)[35], Canonical interval forest (CIF)[36], and Diverse Representation Canonical Interval Forest (DrCIF)[37] are examples of these algorithms. Table 2 shows these algorithms, their main FE algorithms, and their respective classification techniques and algorithms.

Table 2. Conventional ML TSC algorithms using interval-based FE techniques.

| Algorithm Name | FE Algorithms | Classification Technique | Classification Algorithm |
|---|---|---|---|
| FBL | HCTSA Features+ greedy feature selection | Statistical | LDA |
| LPS | Subseries regression tree | Distance-based | KNN |
| TSF | Summary Statistics | DT Ensemble | Time-series Tree |
| CIF | Summary Statistics + Catch22 | DT Ensemble | Time-series Tree |
| DrCIF | Summary Statistics + Catch22 | DT Ensemble | DT |
| TSBF | Summary Statistics + BOP | DT Ensemble | Random Forest DT |
| RISE | Spectral features | DT Ensemble | Random Forest DT |

*Shapelet-based algorithms* use time-series shapelet discovery and shapelet transforms to perform the FE. Shapelets are time-series subsequences that have discriminatory information about class membership. The shapelet algorithm was first proposed by Ye and Keogh [38]. The extracted shapelets are then used to train a classifier (e.g., RF or LR ) to do the TSC task. Several algorithms such as Learned Shapelets (LS)[21], Fast Shapelets (FS), Shapelet Transform Classifier (STC)[39], and The Generalised Random Forest (gRFS)[40] are examples of these algorithms. Table 3 shows these algorithms, their main FE algorithms, and their respective classification techniques and algorithms.

Table 3. Conventional ML TSC algorithms using Shapelet-based FE techniques.

| Algorithm Name | FE Algorithms | Classification Technique | Classification Algorithm |
|---|---|---|---|
| LS | KMEANS, Shapelet Discovery | Statistical | LR |
| FS | SAX, Shapelet Discovery | DT Ensemble | DT |
| STC | Shapelet transform | DT Ensemble | Random Forest DT |
| gRFS | Shapelet transform | DT Ensemble | Random Forest DT |

*Differential-based algorithms* are based on the first-order differences of the time-series. Similar to instance-based classification, they use a defined distance metric to compare time-series instances, but they perform this on the transformed differenced time-series. Complexity Invariant Distance (CID)[21], Derivative DTW (DDTW)[21], and Derivative Transform Distance (DTDC)[21] are distance metrics applied on differenced time-series instances. They are then coupled with a one-NN classifier to carry out the TSC task. We denote them as CID-KNN, DTDC-KNN, and DDTW-KNN in this study. Table 4 shows these algorithms, their main FE algorithms, and their respective classification techniques and algorithms.

Table 4. Conventional ML TSC algorithms using Differential-based FE technique.

| Algorithm Name | FE Algorithms | Classification Technique | Classification Algorithm |
|---|---|---|---|
| KNN-CID | first-order differences | Distance-based | KNN |
| KNN-DDTW | first-order differences | Distance-based | KNN |
| KNN-DTDC | first-order differences, cosine transform | Distance-based | KNN |

*Kernel-based algorithms* use different kinds of kernels (e.g., convolutional kernels) to perform the FE task. The extracted features are then used with a classifier (e.g., Ridge classifier) to do the TSC tasks. It should be noted that the weights in convolutional kernels used in this kind of FE are randomly generated and there is no learning process in these kernels. Thus, they are categorized in conventional ML algorithms rather than ANN/DL algorithms. The random convolutional kernel transform (ROCKET)[41], ARSENAL[37], and Time Warping Invariant Echo State Network (TWIESN)[19] algorithms use kernel-based techniques for FE. Table 5 shows these algorithms, their main FE algorithms, and their respective classification techniques and algorithms.

Table 5. Conventional ML TSC algorithms using Kernel-based FE techniques.

| Algorithm Name | FE Algorithms | Classification Technique | Classification Algorithm |
|---|---|---|---|
| ROCKET | convolution kernels | Statistical | Ridge Classifier |
| ARSENAL | convolution kernels | Algorithm Ensemble | ROCKET |
| TWIESN | ESN kernels | Statistical | Ridge Classifier |

*Hybrid algorithms* combine FE techniques from different categories and ensemble the results of multiple TSC algorithms. They are referred to as *meta-ensemble* algorithms. One example is the Collective Of Transformation-based Ensembles (COTE)[19] which is an ensemble of 35 classifiers. COTE has been improved upon with the introduction of a Hierarchical Vote system, new classifiers, and additional representation transformation domains, resulting in HIVE-COTE V1.0[35] and HIVE-COTE V2.0[37]. Another example is the Time Series Combination of Heterogeneous and Integrated Embedding Forest (TS-CHIEF)[42] algorithm which is an ensemble of trees incorporating distance, dictionary, and spectral base features. Table 6 provides details on these algorithms, their main FE algorithms, and their respective classification techniques and algorithms.

Table 6. Conventional ML TSC algorithms using Hybrid FE techniques.

| Algorithm Name | FE Algorithms | Classification Technique | Classification Algorithm |
|---|---|---|---|
| COTE | hybrid | Algorithm Ensemble | EE, ST, ACF, PS |
| HIVE-COTE V1.0 | hybrid | Meta Ensemble | EE Ensemble, Shapelet Ensemble, BOSS Ensemble, TSF, RISE |
| HIVE-COTE V2.0 | hybrid | Meta Ensemble | STC, TDE, ARSENAL, DrCIF |
| TS-CHIEF | similarity measures, dictionary, spectral features | DT Ensemble | Proximity Forest DT |

Two additional algorithms worth mentioning did not fit into any of the major categories mentioned above. The recurrence Plot Compression Distance (RPCD) algorithm uses Recurrence Plots (RP) for time-series transformation coupled with the KNN algorithm to accomplish the TSC task[43]. The Rotation Forest (RotF) algorithm uses a statistical technique (i.e., Principal Component Analysis (PCA)) to extract important features from the raw time-series and then an RF algorithm as the classifier[44]. These two are denoted as Statistical and Graph-based techniques in Figure 4.

Table 7. Instance-based Conventional TSC ML algorithms that do not use any FE techniques.

| Algorithm Name | Classification Technique | Classification Algorithm | Algorithm Name | Classification Technique | Classification Technique |
|---|---|---|---|---|---|
| LR | Statistical | LR | NB | Statistical | NB |
| LDA | Statistical | DA | QDA | Statistical | DA |
| SVM | Distance-based | SVM | BAG-DT | DT Ensemble | DT |
| RF | DT Ensemble | DT | GBM | DT Ensemble | DT |
| Extreme RF | DT Ensemble | RF DT | XGBoost | DT Ensemble | DT |
| KNN-EUC | Distance-based | KNN | KNN-DTW | Distance-based | KNN |
| KNN-LCSS | Distance-based | KNN | KNN-ERP | Distance-based | KNN |
| KNN-WDTW | Distance-based | KNN | KNN-MSM | Distance-based | KNN |
| KNN-TWE | Distance-based | KNN | KNN-DTW-I | Distance-based | KNN |
| KNN-DTW-D | Distance-based | KNN | KNN-DTW-A | Distance-based | KNN |
| EE | Algorithm Ensemble | 11 Elastic metrics | | | |

As mentioned earlier, instance-based approaches are algorithms that do not apply any kind of FE on the data and perform the classification directly on the raw time-series. The major difference between these algorithms is the classification technique they are using and the way they calculate the distance or measure similarity between different time-series instances. In our study, we found 23 of these instances. Algorithms such as Naive Bayes (NB), Quadratic Discriminant Analysis (QDA), BAG-DT, Gradient Boosting Machine (GBM), Extreme RF, eXtreme Gradient Boosting (XGBoost)[45], Proximity Forest (PF)[46], and one Nearest Neighbor with different elastic distance metrics such as Euclidean, Dynamic Time Warping (DTW)[21], Longest Common SubSequence (LCSS)[21], Edit distance with Real Penalty (ERP)[21], Weighted DTW[21], Move-Split-Merge (MSM)[21], Time Warp Edit (TWE)[21] are examples of these algorithms. In this study, we denote these algorithms with the classifier name followed by the distance metric. Moreover, Independent, Dependent, and Adaptive KNN-DTW are the generalized version of DTW adopted for multivariate time-series[47], and Elastic Ensemble (EE) is an ensemble algorithm that works by ensembling eleven elastic metrics[21]. Table 7 shows these algorithms, and their respective classification techniques and algorithms.

### 2.1.1.2 Classification Techniques

Conversely, we can look at the different algorithms and categorize them based on their classification technique. We distinguish five different classification techniques. Figure 5 depicts these techniques that are used in literature for conventional ML TSC algorithms. More details on each algorithm can be found in the referenced papers in subsequent sections.

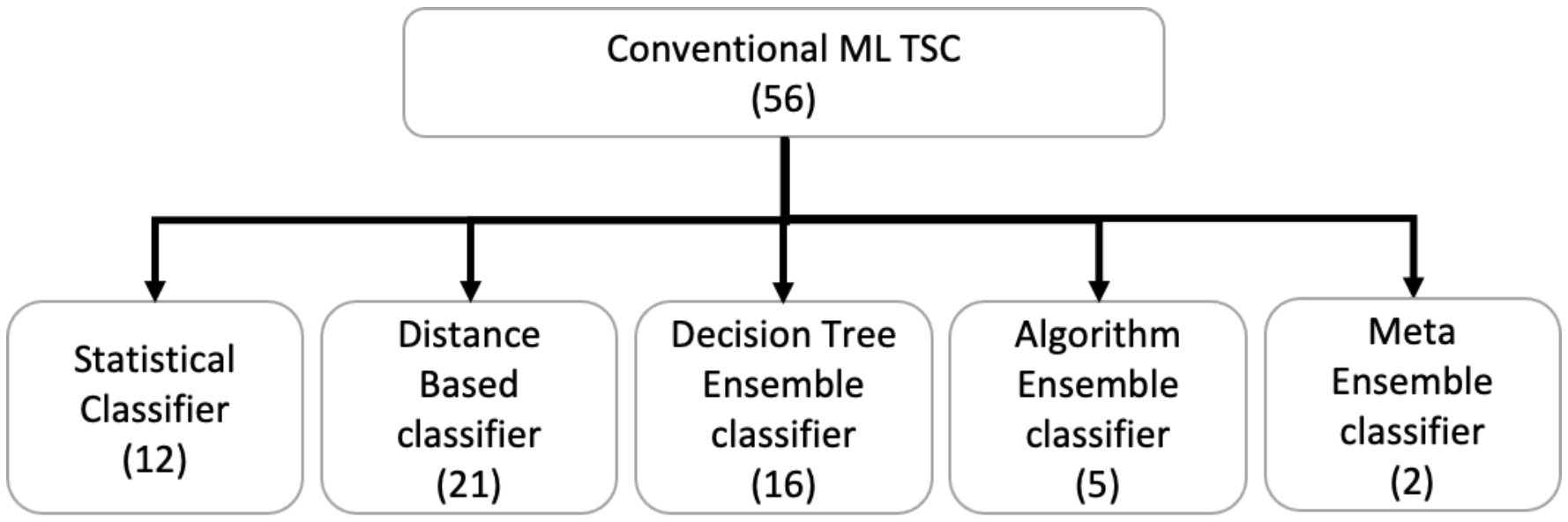

Figure 5: Classification techniques in conventional ML TSC algorithms

*Statistical classifiers* use statistical classification algorithms (e.g., LR, NB, DA, and Ridge classifiers) and techniques (e.g., probability or maximum likelihood) to calculate the distance or measure similarity between different time-series instances or extracted features from the FE module. Algorithms such as LR, NB, LDA, QDA, FBL, LS, WEASEL+MUSE, MrSEQL, MrSQM, TWIESN, and ROCKET algorithms use statistical classifiers to carry out the TSC tasks. Table 8 shows these algorithms and their respective FE techniques and algorithms.

Table 8. Conventional ML TSC algorithms using Statistical classification techniques.

| Algorithm Name | Classification Algorithms | FE Technique | FE Algorithm |
|---|---|---|---|
| LR | LR | None | None |
| NB | NB | None | None |
| DA | DA | None | None |
| QDA | DA | None | None |
| Ridge Classifier | Ridge | None | None |
| FBL | LDA | Interval-based | HCTSA Features+ greedy feature selection |
| LS | LR | Shapelet-based | k-means clustering, Shapelet discovery |
| WEASEL+MUSE | LR | Dictionary-based | SFA, first order differences-SFA |
| MrSEQL | LR | Dictionary-based | SFA, SAX |
| MrSQM | LR | Dictionary-based | SFA, SAX |
| ROCKET | Ridge Classifier | Kernel-based | Convolution Kernels |
| TWIESN | Ridge Classifier | Kernel-based | ESN kernels |

*Distance-based classifiers* address the TSC tasks by defining a distance metric that measures the distance between the sequential values directly or between extracted features from the FE module. The defined distance metric is then used inside a KNN for all these algorithms except DTWF and SVM which use SVM as the classifier. Euclidean distance, DTW, WDTW, TWE, MSM, LCSS, ERP, CID, DDTW, and DTDC are different distance metrics coupled with a 1-NN classification algorithm. We denote these algorithms with the classifier name followed by the distance metric. The details of these algorithms can be found in the great work done by Bagnall et al.[21]. There are three strategies for using DTW for multivariate problems, proposed by Shokoohi-Yekta et al. [47]. KNN-DTW-I considers different dimensions in a multivariate time-series to be independent. KNN-DTW-D considers them to be dependent and KNN-DTW-A considers an adaptive approach. LPS, BOP, SAXVSM, RPCD, BOSSVS, and BOSS algorithms are all distance-based TSC algorithms that use the KNN classifier and DTWF uses the SVM algorithm on extracted features techniques to do the TSC task. Table 9 shows these algorithms, and their respective FE techniques and algorithms.

Table 9. Conventional ML TSC algorithms using Distance-based classification techniques.

| Algorithm Name | FE Technique | FE Algorithms | Algorithm Name | FE Technique | FE Algorithms |
|---|---|---|---|---|---|
| KNN-EUC | None | None | KNN-DTDC | Differential-based | First-order differences, cosine transform |
| KNN-DTW | None | None | SAXVSM | Dictionary-based | SAX-VSM |
| SVM | None | None | RPCD | Graph-based | Recurrence Plots |
| KNN-LCSS | None | None | KNN-TWE | None | None |
| KNN-ERP | None | None | BOSS | Dictionary-based | SFA |
| KNN-WDTW | None | None | BOSSVS | Dictionary-based | SFA, tf-idf |
| BOP | Dictionary-based | PAA, SAX | DTWF | Dictionary-based | DTW, SAX |
| KNN-MSM | None | None | LPS | Interval-based | Subseries regression tree |
| KNN-CID | Differential-based | first-order differences | KNN-DTW-I | None | None |
| KNN-DDTW | Differential-based | first-order differences | KNN-DTW-D | None | None |
| KNN-DTW-A | None | None | | | |

*Decision tree Ensemble classifiers* are a group of classifiers that utilize ensemble learning techniques for classification. They do the ensemble learning on the results generated by multiple decision tree-based classification algorithms (e.g., Decision Tree, Random Forest, and Proximity Forest). The ensemble learning can be implemented on either the raw time-series or on the features extracted in the FE module. In our review, BAG-DT[48], RF[49], GBM, Extreme RF, XGBoost[45], and PF[46] algorithms use decision tree ensemble techniques on raw time-series for classification. Moreover, RotF[44], FS, TSF, TSBF, gRFS, RISE, CIF, DrCIF, STC, and TS-CHIEF algorithms use decision tree ensemble techniques on the features extracted by a FE module for classification. Table 10 shows these algorithms, and their respective FE techniques and algorithms.

Table 10. Conventional ML TSC algorithms using DT Ensemble classification techniques.

| Algorithm Name | Classification Algorithms | FE Technique | FE Algorithm |
|---|---|---|---|
| BAG-DT | DT | None | None |
| RF | DT | None | None |
| GBM | DT | None | None |
| Extreme RF | Random Forest DT | None | None |
| RotF | Random Forest DT | Statistical | PCA |
| FS | DT | Shapelet-based | SAX, Shapelet Discovery |
| TSF | Time-series Tree | Interval-based | Summary Statistics |
| CIF | Time-series Tree | Interval-based | Summary Statistics + Catch22 |
| DrCIF | DT | Interval-based | Summary Statistics + Catch22 |
| TSBF | Random Forest DT | Interval-based | Summary Statistics +BOP |
| STC | Random Forest DT | Shapelet-based | Shapelet transform |
| XGBoost | DT | None | None |
| gRFS | Random Forest DT | Shapelet-based | Shapelet transform |
| RISE | Random Forest DT | Interval-based | Spectral features |
| PF | Proximity Forest DT | None | None |
| TS-CHIEF | Proximity Forest DT | Hybrid | Similarity measures, dictionary representations, interval-based transformations |

*Algorithm Ensemble classifiers* are similar to DT ensemble classifiers in the sense that they also use ensemble learning. The difference is the classifiers in this group do the ensemble learning on the results of multiple different algorithms other than DTs. EE, COTE, CBOSS, TDE, and ARSENAL algorithms use algorithm-ensemble techniques for classification. Table 11 shows these algorithms, and their respective FE techniques and algorithms.

Table 11. Conventional ML TSC algorithms using Algorithm Ensemble classification techniques.

| Algorithm Name | Classification Algorithms | FE Technique | FE Algorithm |
|---|---|---|---|
| EE | 11 Elastic metrics | None | None |
| COTE | EE, ST, ACF, PS | Hybrid | Hybrid |
| CBOSS | BOSS | Dictionary-based | SFA |
| TDE | BOSS | Dictionary-based | SFA |
| ARSENAL | ROCKET | Kernel-based | Convolution kernels |

Finally, *Meta Ensemble classifiers* are groups of classifiers that do the ensemble learning on a group of algorithm ensemble classifiers. These algorithms are considered the state of the art for TSC tasks but they become hugely computationally intensive and impractical to run on a real big problem[19]. HIVE-COTE V1.0 and HIVE-COTE V2.0 algorithms use meta-ensemble techniques for classification. Table 12 shows these algorithms, and their respective FE techniques and algorithms.

Table 12. Conventional ML TSC algorithms using Meta Ensemble classification techniques.

| Algorithm Name | Classification Algorithms | FE Technique | FE Algorithm |
|---|---|---|---|
| HIVE-COTE V1.0 | EE Ensemble, Shapelet Ensemble, BOSS Ensemble, TSF, RISE | Hybrid | Hybrid |
| HIVE-COTE V2.0 | STC, TDE, ARSENAL, DrCIF | Hybrid | Hybrid |

### 2.1.2. ANN and DL TSC Algorithms

Over the past years and in line with researchers' access to more and cheaper computational power, ANN and DL algorithms have had great success in various fields for classification tasks. In 2015, deep Convolutional Neural Networks (CNNs) revolutionized the field of computer vision by reaching human-level accuracy[19]. Following the success of DL in computer vision, many researchers started proposing Deep Neural Networks (DNN) architectures to solve problems in other domains, such as Natural Language Processing (NLP), Neural Image Compression[50,51], and speech recognition. Different ANN and DL architectures can be used for TSC tasks[52]. The main difference between these architectures is mainly in the FE module of the overall classification algorithm. There are several techniques and architectures that are used in ANN/DL TSC algorithms. In this study, we explored 36 algorithms divided into five categories. All algorithms that we explored use the same classification technique (i.e., statistical) and classification algorithm and function (i.e., Sigmoid for binary classification and Softmax for multiclass classification) so it will not be repeated for the rest of the paper. Figure 6 shows the architectures that are used in the focal ANN/DL TSC algorithms. The numbers in parentheses correspond to the count of algorithms in each category from the initial compiled list of 92 algorithms. More details on each algorithm can be found in the referenced papers in subsequent sections.

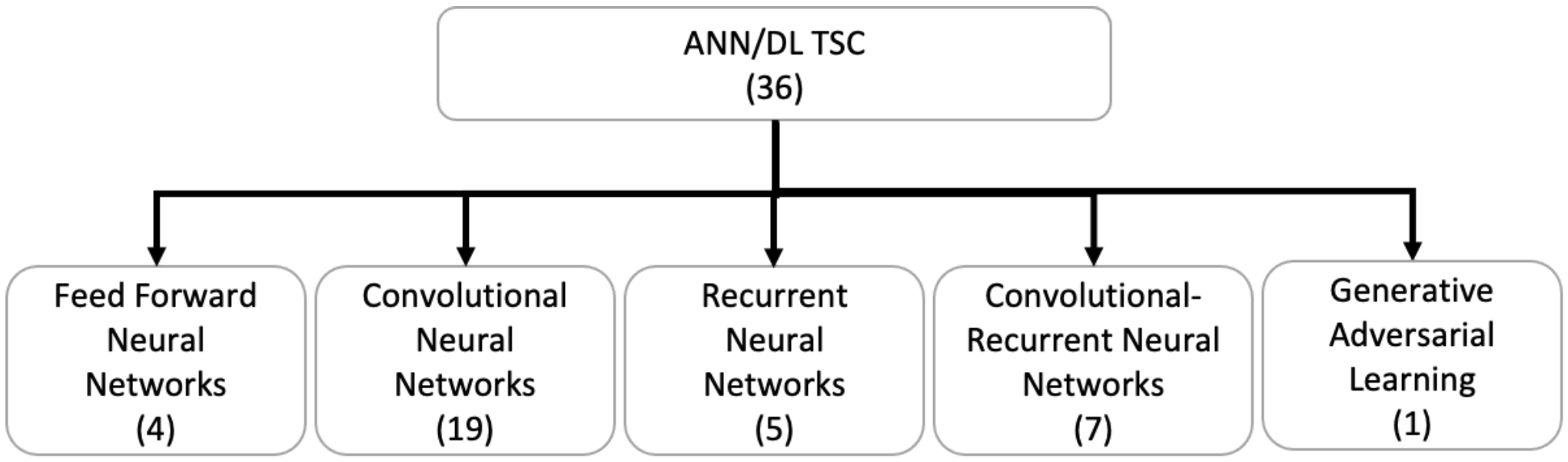

Figure 6: Different architectures used in ANN/DL TSC algorithms

The *Feed Forward Neural Networks (FFN)* are the first and simplest type of ANN architecture that was proposed. The main characteristic of these networks is that the information moves in only one direction and there are no cycles or loops in the network. Multilayer Perceptron (MLP)[53], FFT-MLP[54], Ensemble Sparse Supervised Model (ESSM)[55], and DA-NET[17] algorithms are using the FFN-based architecture to do the TSC tasks. Table 13 shows these algorithms, and their respective FE architectures and algorithms.

Table 13. ANN & DL ML TSC algorithms using FFN feature engineering architectures.

| Algorithm Name | FE Technique | FE Layers |
|---|---|---|
| MLP | FFN | MLP |
| FFT-MLP | FFN | FFT, MLP |
| DA-NET | FFN | MLP, Dual Attention (SEWA & SSAW), MLP |
| ESSM | FFN | Sparse filtering, MLP |

*Convolutional Neural Networks (CNN)* are another group of ANN that was originally proposed by LeCun in 1998 for image analysis. Since then, many variations of CNN have been proposed and successfully applied for different tasks. CNN architectures have shown good results in time series classification due to their powerful local feature capture capabilities. The main characteristic of CNN is using convolution kernels with trainable weights to find the local patterns by high-dimensional nonlinear feature extraction[17]. CNN[54], Time-CNN[19], FDC-CNN[56], Fully Convolutional Networks (FCN)[53], t-LeNet[19], HHO-ConvNet[57], 1DCNN[58,59], Encoder[19], MCDCNN[19], Multiple Time-Series Convolutional Neural Network (MTS-CNN)[20], MCNN[19], Dilated CNN[60], ResNet[23,53], Temporal Convolutional Networks (TCN)[61], InceptionTime[62], Inception-1DCNN[63], MultiVariate Convolutional Neural Network (MVCNN)[64], CWT-CNN[65,66], GASF-CNN[67] algorithms are using different variants of CNN-based for the TSC tasks. It is important to note that here we are only exploring the general architecture of these algorithms and different possibilities for each category of algorithms. Although several algorithms may seem similar to each other, the fine details of each of them such as the number of layers, number of kernels in each layer, etc. are different. These details are out of the scope of this study and we encourage interested readers to find them in referenced papers for each algorithm. Table 14 shows these algorithms, and their respective FE architectures and algorithms.

Table 14. ANN & DL ML TSC algorithms using CNN FE architectures.

| Algorithm Name | FE layers | Algorithm Name | FE layers |
|---|---|---|---|
| CNN | Conv1D--MLP | MCNN | Window Slicing, Conv1D, MLP |
| Time-CNN | Conv1D, MLP | Dilated CNN | Dilated Conv, Conv1D Layers, MLP Layers |
| FDC-CNN | Conv1D, MLP | ResNet | Conv1D, Residual Block, MLP |
| FCN | Conv1D, MLP | TCN | dilated causal convolution, Residual block, MLP |
| t-LeNet | Conv1D, MLP | InceptionTime | Conv1D, Inception Block, Residual Block, MLP |
| HHO-ConvNet | Conv1D, MLP | Inception-1DCNN | Inception Conv, Conv1D, MLP |
| 1DCNN | Conv1D, MLP | MVCNN | 1*1 Conv, Inception Conv, MLP |
| Encoder | Conv1D, Attention, MLP | CWT-CNN | CWT, Conv2D, MLP |
| MCDCNN | Independent Conv1D on channels, MLP | GASF-CNN | PAA, GASF, Conv2D, MLP |
| MTS-CNN | Independent Conv1D on channels, Independent MLP on channels, MLP | | |

*Recurrent Neural Networks (RNN)* are another popular type of architecture for ANN/DL algorithms for TSC tasks. RNNs have been developed to address the sequential input data and have the sequential data feeding ability. The main characteristic of these networks is their memory as they take information from prior time stamps to influence the current input and output. While other ANN architectures assume that the input variables are independent of each other, the output of an RNN depends on the prior elements within the sequence. In our review, RNN[68], LSTM[61], Stacked LSTM[69], BI-LSTM[70], and TS-LSTM[65] algorithms use different variants of RNN-based architectures for the TSC tasks. Table 15 shows these algorithms, and their respective FE architectures and algorithms.

Table 15. ANN & DL ML TSC algorithms using FFN FE architectures.

| Algorithm Name | FE Technique | FE Layers |
|---|---|---|
| RNN | RNN | RNN--MLP |
| LSTM | RNN | LSTM--MLP |
| Stacked LSTM | RNN | LSTM, MLP |
| BI-LSTM | RNN | Bi-LSTM, MLP |
| TS-LSTM | RNN | LSTM, Attention, MLP |

Some algorithms utilize a combination of both RNN and CNN layers inside their architectures. This is to make use of both of these network's characteristics and to increase the performance of the TSC algorithm overall. Since this group cannot fit neatly in either of those categories, we denote these algorithms as *CNN-RNN architectures*. In our review, ATT-1DCNN-GRU[71], CNN-LSTM[72], 4-layer CNN-LSTM[69], Time-series attentional prototype network (TapNet)[73], Multivariate LSTM-FCN (MLSTM-FCN)[74,75], Multivariate Attention LSTM-FCN (MALSTM-FCN)[75], FFT-CNN-LSTM[76] algorithms are using different variants of CNN-RNN based architectures to the TSC task. Table 16 shows these algorithms, and their respective FE architectures and algorithms.

Table 16. ANN & DL ML TSC algorithms using FFN FE architectures.

| Algorithm Name | FE Technique | FE Layers |
|---|---|---|
| ATT-1DCNN-GRU | CNN-RNN | Conv1D--GRU--Attention--MLP |
| CNN-LSTM | CNN-RNN | Conv1D, LSTM |
| 4-layer CNN-LSTM | CNN-RNN | Conv1D, LSTM, MLP |
| TapNet | CNN-RNN | Conv1D, LSTM, MLP, Attention |
| MLSTM-FCN | CNN-RNN | LSTM, Conv1D, Squeeze & Excitation |
| MALSTM-FCN | CNN-RNN | LSTM, Attention, Conv1D, Squeeze & Excitation |
| FFT-CNN-LSTM | CNN-RNN | FFT, Conv2D, LSTM, MLP |

Finally, a more recent ANN architecture is *Generative Adversarial Networks (GAN)*. It was developed by Goodfellow et al.[77]. The idea behind GAN is to make use of two different kinds of networks (i.e., discriminator and generative) and make them compete with each other to learn the joint probability between a set of input features and output classes[78]. Based on our review, the use of GAN neural networks for TSC tasks in manufacturing has been very limited, and we only had the WGAN-GP-based deep adversarial transfer learning (WDATL) algorithm that used this architecture[79].

## 2.2. Public Manufacturing TSC Datasets

One of the greatest challenges when studying time-series analytics for smart manufacturing applications is the availability of applicable public datasets[10]. Currently, there are a limited number of preprocessed manufacturing datasets that are publicly available for researchers and practitioners. As a result, ML researchers in the manufacturing domain either have to i) revert to primary data collected from machines which is hard to come by, ii) do the preprocessing from scratch based on their application, or iii) turn to popular datasets found in other domains. While valuable, these datasets from other domains lack characteristics that are unique to and representative of manufacturing problems. There are many publicly available time-series datasets covering a range of applications outside of manufacturing, such as in the medical domain, the financial markets, human activity recognition, and speech recognition applications. The University of California Riverside (UCR) time-series archive[80], and the University of East Anglia (UEA) multivariate time-series classification archive[81], are the main examples of these publicly available datasets. Additionally, there are also some time-series datasets available at the UC Irvine (UCI) Machine Learning repository[2]. Each of these sources contains preprocessed datasets that can be used for research purposes. However, the number of manufacturing-related public datasets that can be used in TSC is limited and covers a narrow range of applications. A persuasive reason for this scarcity of manufacturing datasets is the reluctance of some authors to share their datasets and non-disclosure agreements with the industry in many cases. As a result, manufacturing is missing out on valuable advantages afforded to the CS field by strong data availability to advance the field as a whole.

In this study, we aim to bridge this gap by gathering several available datasets from various manufacturing resources and preprocessing them with a standard and transparent methodology. The result is a repository of ready-to-use manufacturing-specific datasets that can be fed into ML algorithms to investigate their performance in a smart manufacturing setting.

Table 17. The initial list of 33 manufacturing datasets with different applications used for a variety of tasks

| Dataset Name | Domain | Associated Application | ML Task | Reference |
|---|---|---|---|---|
| Gas Sensor Temperature | Semiconductor | Detection limit Estimation | Forecasting | UCI ML Repository |

[2] https://archive.ics.uci.edu/ml/index.php

| | | | | |
|---|---|---|---|---|
| Hydraulic systems | Railway | Condition Monitoring | Classification | UCI ML Repository |
| Gas sensors home activity | Chemical | Condition Diagnosis | Classification | UCI ML Repository |
| Control charts Time-series | General | Pattern Recognition | Classification | UCI ML Repository |
| PHM09 Gearbox | Gearbox | Anomaly Detection & Prognosis | Anomaly Detection, Forecasting | PHM Society |
| PHM10 CNC milling | Milling | RUL Estimation | Forecasting | PHM Society |
| PHM18 Ion Mill Etch | Semiconductor | RUL Estimation | Forecasting | PHM Society |
| PHM22_Rock_Drills | Rock drills | Fault Diagnosis | Classification | PHM Society |
| Paderborn University | Bearing | Fault Diagnosis | Classification | Neupane et.al.,2020[5] |
| NASA FEMTO | Bearing | RUL Estimation | Forecasting | Neupane et.al.,2020[5] |
| IMS_Bearing | Bearing | Fault Prognosis | Forecasting | Neupane et.al.,2020[5] |
| PHM08 NASA engine | Aerospace | RUL Estimation | Forecasting | PHM Society |
| PHM15 NASA HIRF | Energy | Fault detection and Prognosis | Anomaly Detection, Forecasting | PHM Society |
| PHM19 NASA Crack | Manufacturing | Fault Estimation | Forecasting | PHM Society |
| PHM21 NASA Turbo Fan | Engines | RUL Estimation | Forecasting | PHM Society |
| Bearing_Univar | Bearing | Fault Diagnosis | Classification | Huang et.al., 2013[82] |
| NASA Milling BEST | Milling | Tool wear prognosis | Forecasting | Agogino et. al, 2007[83] |
| NASA MOSFET | Semiconductor | RUL Estimation | Forecasting | Celaya et. al., 2011[84] |
| Battery Dataset | Electronics | RUL Estimation | Forecasting | Saha et. al., 2007[85] |
| NASA IGBT Accelerated | Electronics | RUL Estimation | Forecasting | Celaya et. al., 2009[86] |
| NASA CFRP Composites | Manufacturing | Fault Diagnosis | Classification | Saxena et. al.,[87] |
| 3W | Oil Wells | Anomaly Detection | Anomaly Detection | Vargas et. al., 2019[88] |
| MFPT | Bearing | Fault Diagnosis | Classification | Neupane et.al.,2020[5] |
| Energy Consumption | Energy | Energy Estimation | Forecasting | Data.gov[3] |
| Metal Etching | Semiconductor | Fault Diagnosis | Classification | Wise et. al., 1999[89] |
| CWRU Bearing | Bearing | Fault Diagnosis | Classification | Neupane et.al.,2020[5] |

[3] https://bloomington.data.socrata.com/stories/s/hgqr-8ivd

| | | | | |
|---|---|---|---|---|
| SECOM | Semiconductor | Fault Diagnosis | Classification | UCI ML Repository |
| PHM11 Anemometer | Wind turbines | Anomaly Detection & Fault Prognosis | Anomaly detection, Forecasting | PHM Society |
| PHM13 Maintenance | Unknown | Fault Diagnosis | Classification | PHM Society |
| PHM16 CMP | Semiconductor | MMR Prediction | Forecasting | PHM Society |
| PHM17 Train Bogie | Transportation | Fault Diagnosis | Anomaly Detection | PHM Society |
| NASA IMS Bearing | Bearing | Fault Diagnosis | Classification | Lee et. al., 2007[90] |
| Metal Etching Feature-set | Semiconductor | Fault Diagnosis | Classification | Wise et. al., 1999[89] |

The goal of this study is to perform the experimental evaluation of previously mentioned TSC algorithms on manufacturing-related datasets and evaluate which algorithm(s) are best suited for that situation. To find a representative list of datasets from the manufacturing domain, we first gathered a list of *33 datasets* from various resources. These are manufacturing-related datasets that can be used for different applications and ML tasks. Neupane et al., introduced five datasets that can be used for bearing fault detection and diagnosis applications[5]. The Prognostics and Health Management (PHM) Society is another resource that hosts a data-driven competition every year and has a few public datasets that can be used for time-series research. Jia et al. reviewed these datasets from 2008 until 2017 in their study[91]. The more recent PHM challenge datasets are available on their website[4]. NASA has a Prognostics Center of Excellence Data Set Repository, which is a collection of data sets that have been donated by universities, agencies, or companies for prognosis purposes[5]. Most of these are time-series datasets that fall into the manufacturing domain and can be used for our purpose. It is important to note that this is not an exhaustive list of all available datasets. The datasets with image data types are not included in this selection and we are only focusing on datasets with time-series data type. Table 17 shows our initial list of datasets with their general characteristics.

# 3. Methodology

In this section, we will discuss the methodology we applied to identify and select TSC algorithms for our experiment. Additionally, we will explain the methodology used for manufacturing TSC dataset selection and preprocessing. Furthermore, we will describe the experimental setup used to evaluate the selected algorithms on the manufacturing datasets. It is worth noting that our approach to algorithm selection differs from other references, as we categorized our algorithms based on both FE and classification techniques, whereas other references solely categorized their algorithms based on the FE techniques they used.

[4] https://phmsociety.org/
[5] https://www.nasa.gov/content/prognostics-center-of-excellence-data-set-repository

## 3.1. Algorithms

After compiling an initial list of *92 TSC algorithms* and categorizing them based on FE and classification techniques, we developed a methodology to select a representative list of algorithms. The final selection considered defined requirements and parameters.

*Conventional ML algorithms* were categorized *based on the classification technique* they used (refer to Figure 5). Within each classification category, we sorted the algorithms by *their proposed year* and further categorized them based on *the FE technique* employed. We assumed that newer algorithms within the same category are more evolved and can represent that group. This assumption was made due to the limited time and resources to run all 92 algorithms. Hence, we selected the most modern algorithm from each group as a starting point as the most advanced of the group, with the intention of comparing other algorithms to the selected ones in future works.

To illustrate, the RF algorithm proposed in 2001 utilizes a DT ensemble classification technique. In 2013, the TSF algorithm was introduced for TSC tasks, incorporating an interval-based FE module and using RF as the classifier. The CIF algorithm, proposed in 2020, is an evolution of TSF that improved the FE module by adding more representative features. In 2021, the DrCIF algorithm was introduced as an extension to CIF[37]. Therefore, according to our methodology, we will only employ the DrCIF algorithm in our experimental evaluation. Figures 7 to 10 demonstrate our algorithm selection mechanism.

In these figures, the golden cells with bold font represent the benchmark algorithm that we used within each group. Horizontal dashed lines separate different FE techniques. Algorithm groups are visually differentiated by different colors, indicating that algorithms with the same color belong to the same group. Straight line arrows depict the evolutionary progression of algorithms over time, while the selected representative algorithm is indicated with a star at the upper left corner. Within the Meta-ensemble classification group, due to the high number of algorithms, numeric and alphabetic indicators were used within the ensemble algorithms to enhance visibility (Figure 10).

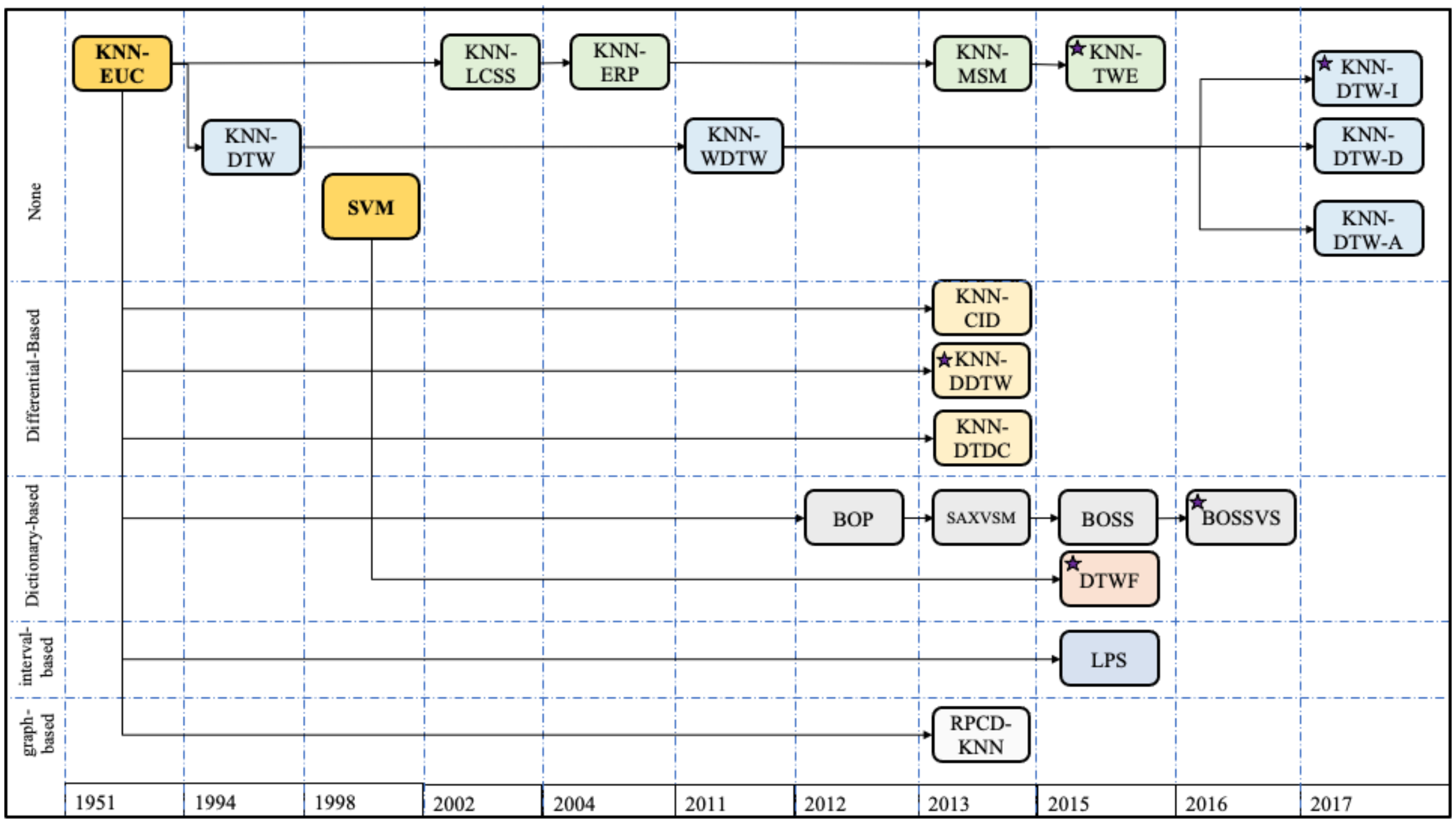

Figure 7: Genealogy of Distance-based classification algorithms

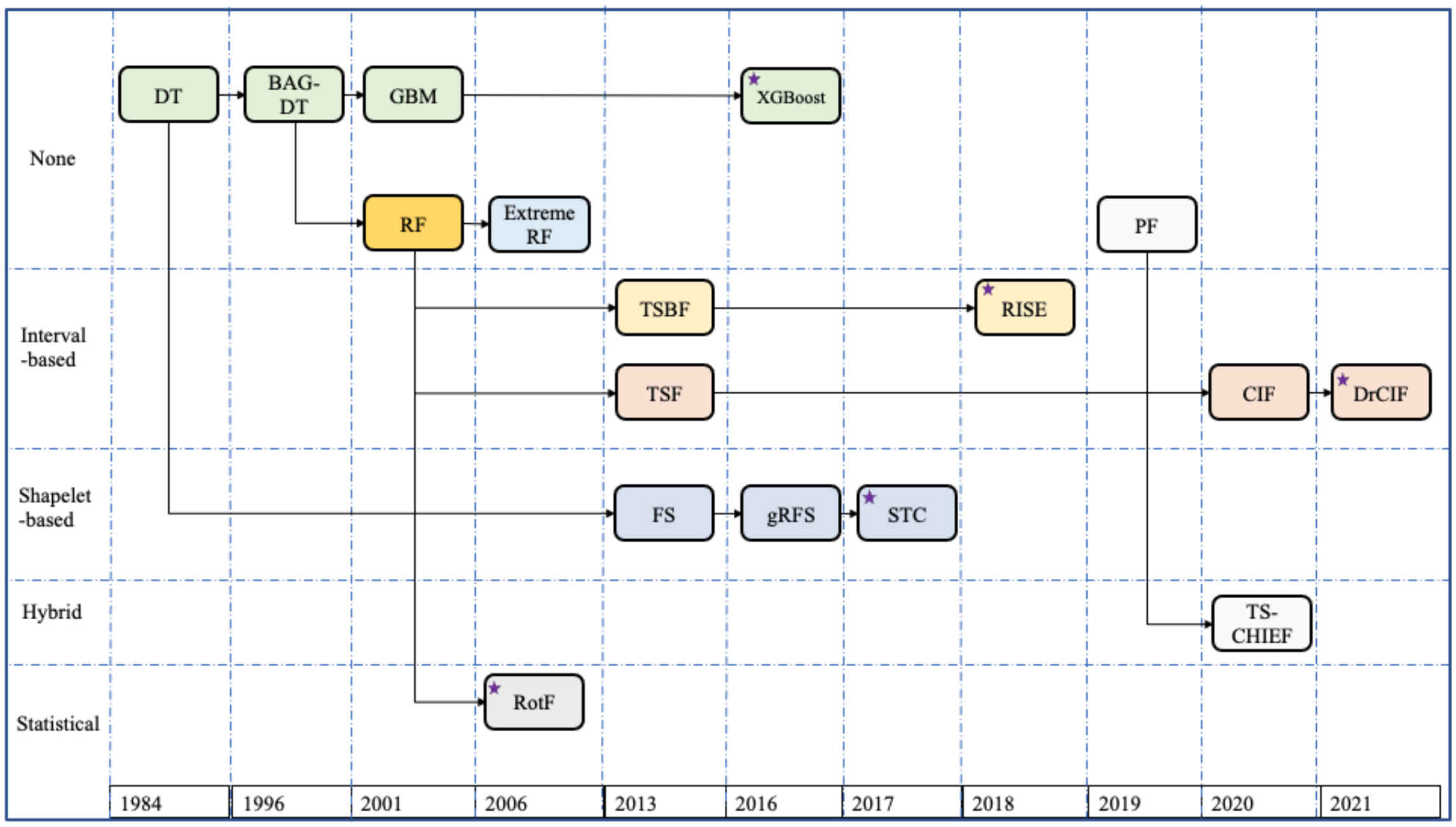

Figure 8: Genealogy of Decision Tree-based classification algorithms

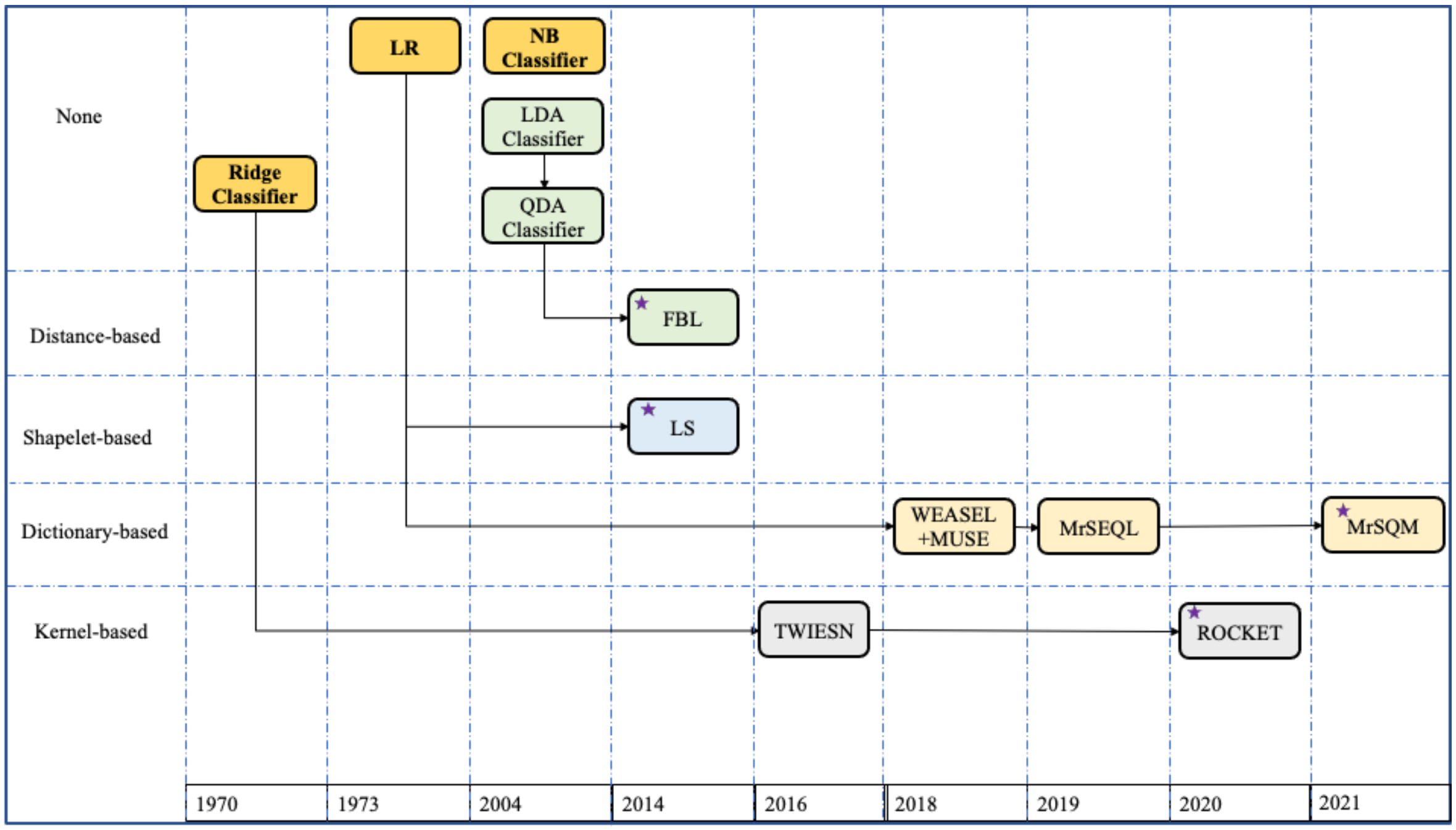

Figure 9: Genealogy of Statistical-based classification algorithms

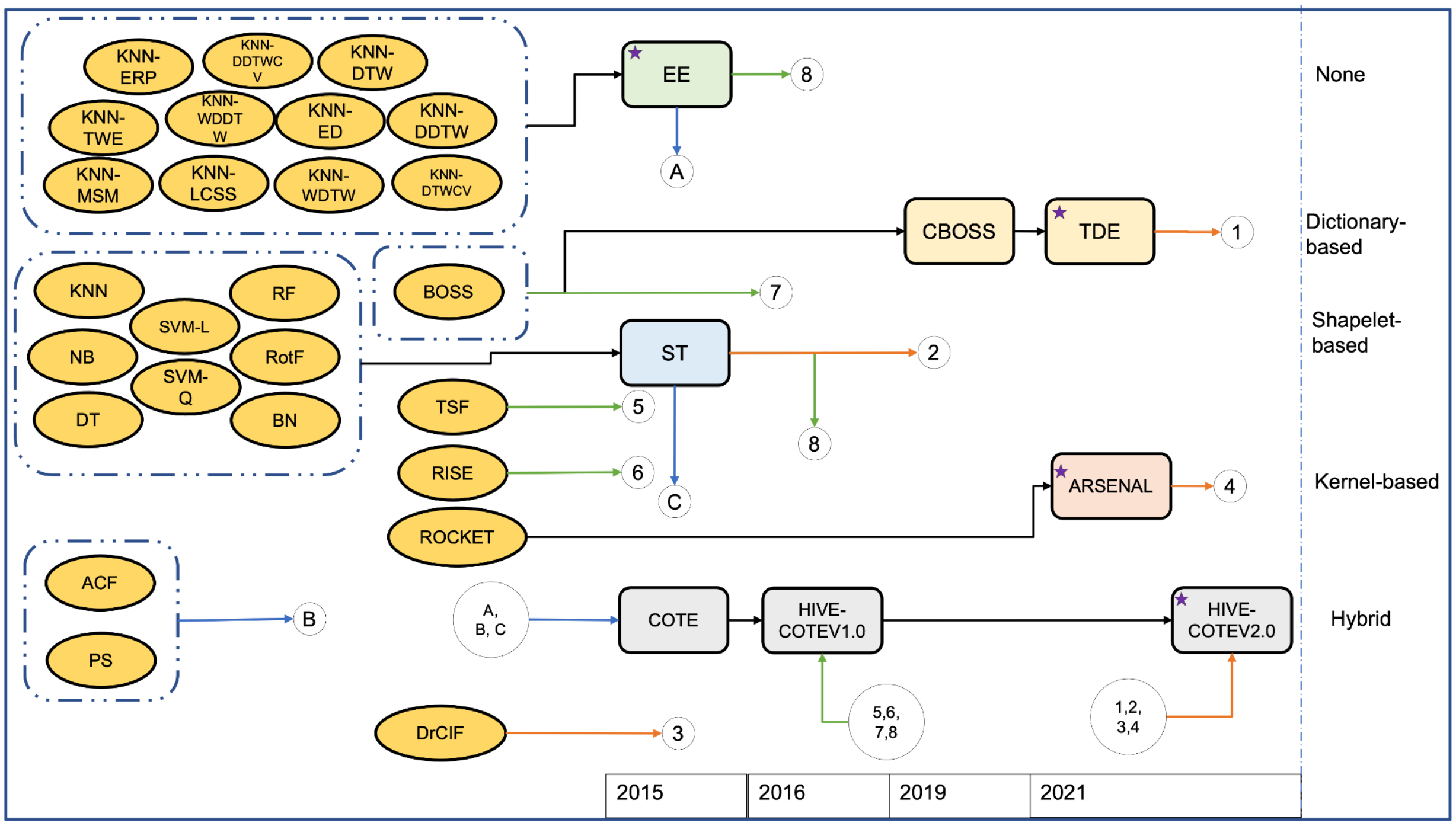

Figure 10: Algorithm ensemble and Meta ensemble classification algorithms

After implementing the methodology and finding representative algorithms, we could not find RPCD-KNN, TS-CHIEF, and LPS algorithms code in our experimental setup environment (i.e., Python). That restriction led to the removal of those algorithms from the final algorithm list.

Table 18. The final set of 19 Conventional ML TSC algorithms selected for experimental evaluation

| Algorithm Name | FE Technique | Classification Technique | Year Proposed | Reference |
|---|---|---|---|---|
| KNN-TWE | None | Distance-based | 2015 | Lines, J., & Bagnall, A.[92] |
| PF | None | DT ensemble | 2019 | Lucas, B., et al.[46] |
| KNN-DTW-I | None | Distance-based | 2017 | Shokoohi-Yekta, M., et al.[47] |
| EE | None | Algorithm Ensemble | 2015 | Lines, J., & Bagnall, A.[92] |
| XGBoost | None | DT ensemble | 2016 | Chen, T., & Guestrin, C.[45] |
| KNN-DDTW | Differential-based | Distance-based | 2013 | Gorecki and Luczak[93] |
| FBL | Distance-based | Statistical | 2014 | B. Fulcher and N. Jones[22] |
| BOSS-VS | Dictionary-based | Distance-based | 2016 | Schäfer, P.[28] |
| DTWF | Dictionary-based | Distance-based | 2015 | Kate, R. J.[29] |
| TDE | Dictionary-based | Algorithm Ensemble | 2021 | Middlehurst, M. et al.[37] |
| MrSQM | Dictionary-based | Statistical | 2021 | Nguyen, T. L., & Ifrim, G.[33] |
| LS | Shapelet-based | Statistical | 2014 | Grabocka et al.[94] |
| STC | Shapelet-based | DT ensemble | 2017 | Bostrom, A., & Bagnall, A.[39] |
| DrCIF | Interval-based | DT ensemble | 2021 | Middlehurst, M. et al.[37] |
| RISE | Interval-based | DT ensemble | 2018 | Lines, J., et al.[35] |
| RotF | Statistical | DT ensemble | 2006 | Rodriguez, J. J., et al.[44] |
| ARSENAL | Kernel-based | Algorithm Ensemble | 2021 | Middlehurst, M. et al.[37] |
| ROCKET | Kernel-based | Statistical | 2020 | Dempster, A., et al.[41] |
| HIVE-COTE 2 | Hybrid | Meta Ensemble | 2021 | Middlehurst, M. et al.[37] |

We were not able to apply the exact same methodology to *ANN/DL algorithms*. The main difference between different *ANN and DL classification algorithms* is in the FE module of the overall classification algorithm. Thus, we used this as the main criterion to categorize the algorithms (see Figure 6). Another criterion we considered for this group of algorithms is the availability of their *algorithm codes in open-source repositories* as a practical limitation. We also looked at algorithms with similar structures and chose *the latest one with available code* as the representative for a category. For instance, the CNN, Time-CNN, FDC-CNN, FCN, t-LeNet, HHO-ConvNet, and 1DCNN algorithms share the same structure of coupling Conv1D and MLP layers with minor differences. Thus, *FCN*[53] was selected for our experimental evaluation to assess the performance of this structure in TSC tasks. Apart from FCN, four other algorithms from the *CNN-based TSC algorithms* group were chosen. The *Encoder*[19] algorithm investigates the impact of the attention mechanism, the *ResNet*[53] algorithm examines the effect of residual blocks, and the *InceptionTime*[62] algorithm assesses the effect of both inception and residual modules.

Four algorithms have been chosen from the *RNN-based TSC algorithms* group. The *Stacked LSTM*[69] algorithm is the baseline representative of RNN-based algorithms, The *BI-LSTM*[70] algorithm evaluates the effect of bidirectional connections in LSTM layers, and finally, the *TS-LSTM*[65] algorithm evaluates the effect of attention mechanism in RNN-based architectures.

From the *FFN-based TSC algorithms* group, the *MLP*[62] algorithm has been chosen as a benchmark for all classification tasks and the *DA-NET*[17] algorithm evaluates transformer-like structures with forward connections for TSC tasks. One other algorithm has been chosen as representative of the *CNN-RNN-based TSC algorithms* group. The *MALSTM-FCN*[75] algorithm evaluates a more sophisticated RNN-CNN structure coupled with attention and squeeze and excitation mechanisms for TSC tasks.

Table 19. The final set of 10 ANN & DL ML TSC algorithms selected for experimental evaluation

| Algorithm Name | FE Technique | Classification Technique | Year Proposed | Reference |
|---|---|---|---|---|
| FCN | CNN | Statistical | 2017 | Wang, Z., et al.[53] |
| Encoder | CNN | Statistical | 2018 | Serrà, J., et al.[95] |
| ResNet | CNN | Statistical | 2017 | Wang, Z., et al.[53] |
| InceptionTime | CNN | Statistical | 2020 | Ismail Fawaz, H., et al.[62] |
| Stacked LSTM | RNN | Statistical | 2021 | Mekruksavanich, S., & Jitpattanakul, A.[69] |
| BI-LSTM | RNN | Statistical | 2021 | Bartosik, S. C., & Amirlatifi, A.[70] |
| TS-LSTM | RNN | Statistical | 2021 | Lee, W. J., et al.[65] |
| DA-NET | FFN | Statistical | 2022 | Chen, R., et al.[17] |
| MALSTM-FCN | CNN, RNN | Statistical | 2019 | Karim, et al[75] |
| GASF-CNN | GASF, CNN | Statistical | 2019 | Martínez-Arellano, et al[67] |

Table 18, Table 19, and Table 20 depict the *final list of selected conventional ML, ANN & DL, and benchmark algorithms* for this experiment. In total, a set of *36 algorithms* including *18 conventional ML algorithms*, *ten DL algorithms*, and *seven benchmark algorithms* have been chosen for this experimental evaluation. It should be noted that the proposed year and reference in these tables are related to the specific implementation of the algorithm we are using.

Table 20. The final set of seven Benchmark ML algorithms selected for experimental evaluation

| Algorithm Name | FE Technique | Classification Technique | Year Proposed | Reference |
|---|---|---|---|---|
| LR | None | Statistical | 1958 | Cox, D. R.[96] |
| NB | None | Statistical | 2004 | H. Zhang[97] |
| RF | None | DT ensemble | 2001 | Breiman, L.[98] |
| SVM | None | Distance-based | 1998 | Cortes, C., & Vapnik, V. [99] |
| KNN-EUC | None | Distance-based | 1951 | Fix & Hodges [100] |
| Ridge | None | Statistical | 1970 | Hoerl, A. E., & Kennard, R. W.[101] |
| MLP | FFN | Statistical | 2017 | Wang, Z., et al.[53] |

## 3.2. Datasets

After compiling the initial list of datasets, the next step is to select a representative list of comparable datasets to use in our experimental evaluation. During the stage of gathering manufacturing-related datasets from various resources, two insights emerged that were noteworthy.

First, we faced *two main types of time-series datasets*. Some datasets provide the raw signals from the sensor readings, which we refer to as *"Raw Time-Series"*. And some others provide multiple features extracted from the time-series signals (e.g., mean, standard deviation, min, max, etc.). We refer to this type of dataset as a *"Feature-set"*. The main difference between these two kinds of datasets is the fact that the consecutive data points in a given sample of the former type are dependent on each other, and there might be some degree of autocorrelation between them. This is not true in the latter case, and consecutive data points can be considered independent. Since TSC on raw time-series is considered a more challenging task, we chose raw time series datasets for this review. From the list in Table 17, SECOM, PHM11 Anemometer, PHM13 Maintenance, PHM16 CMP, PHM17 Train Bogie, NASA IMS Bearing, and Metal Etching Feature-set are feature-sets and the rest are raw time-series.

Second, we checked the dataset documentation and recorded the ML tasks that each dataset was designed for. This information is recorded in the "*ML Task*" column of Table 17. For example, the PHM 2021 NASA Turbofan engine is gathered from a run-to-failure experiment and was originally designed for forecasting tasks and RUL estimation applications. Although there can be preprocessing actions available to change this assumption, we only chose the datasets labeled for classification tasks for this review. As a result, *10 datasets* had these conditions and were chosen.

Since the datasets stem from different sources, there are many differences between them. The differences range from different numbers of files and folders to deal with for a sample, to different file types to store the data (e.g., CSV, txt, Matlab, etc.), to varying lengths of time series, to whether the observations were scaled or not after being recorded. The selected datasets must have a standard structure and characteristics to be comparable. Thus, a significant amount of preprocessing work was required to complete this step. To do so, we loaded the dataset into a single file dataset with the structure defined in Section 2.1 (see Figure 2), dealt with varying lengths with a predefined method that seemed logical for a given dataset, and standardized the observations to have a standard normal distribution with a mean equal to zero and unit standard deviation. Then the dataset instances and dimensions were shuffled randomly to avoid any biases in later steps.

A recurring and controversial question to resolve in this kind of research is whether to standardize the datasets or not. Most of the past research in the literature indicates it is advisable to standardize time-series data[23]. The reasoning is threefold: First, if summary measures such as mean and variance can be used to discriminate, then the problem is considered trivial and thus can be solved with simple methods such as thresholding. Second, some algorithms perform standardization internally thus having non-standardized datasets can distort comparisons of algorithms. Finally, some datasets are already standardized (see Table 21), so standardizing the rest allows for less biased comparison.

Table 21. Dataset initial characteristics before preprocessing

| Dataset Name | Balanced | Varying Length | Standardized | # Classes | (N) | (T) | (M) | # Instances |
|---|---|---|---|---|---|---|---|---|
| **BEARING_Uni** | YES | NO | NO | 8 | 2,560 | 8,192 | 1 | 20,971,520 |
| **PHM2022_Multi** | NO | YES | YES | 12 | 53,785 | 556-748 | 3 | 108,011,835 |
| PHM2022_PIN_Uni | NO | YES | YES | 12 | 53,785 | 556-748 | 1 | 36,003,945 |
| PHM2022_PO_Uni | NO | YES | YES | 12 | 53,785 | 556-748 | 1 | 36,003,945 |
| PHM2022_PDIN_Uni | NO | YES | YES | 12 | 53,785 | 556-748 | 1 | 36,003,945 |
| **ETCHING_Multi** | NO | YES | NO | 2 | 128 | 56-112 | 19 | 243,694 |
| **MFPT_48KHZ_Uni** | NO | YES | YES | 3 | 20 | 146484-585936 | 1 | 5,566,392 |
| MFPT_96KHZ_Uni | NO | YES | YES | 3 | 20 | 146484-585936 | 1 | 5,566,392 |
| PADERBORN_64KHZ_Uni | NO | YES | YES | 3 | 2319 | 249940-299038 | 1 | 595,380,325 |
| PADERBORN_4KHZ_Uni | NO | YES | YES | 3 | 2319 | 249940-299038 | 1 | 595,380,325 |
| **PADERBORN_64KHZ_Multi** | NO | YES | YES | 3 | 2319 | 249940-299038 | 3 | 1,786,140,975 |
| PADERBORN_4KHZ_Multi | NO | YES | YES | 3 | 2319 | 16000-299038 | 5 | 1,860,365,981 |
| Hydraulic_sys_10HZ_Multi | NO | YES | NO | 4 | 2205 | 60-6000 | 17 | 96,314,400 |
| **Hydraulic_sys_100HZ_Multi** | NO | YES | NO | 4 | 2205 | 60-6000 | 17 | 96,314,400 |
| **Gas_sensors_home_activity** | YES | YES | NO | 3 | 99 | 3825-15393 | 10 | 928,991 |
| **Control_charts** | YES | NO | NO | 6 | 600 | 60 | 1 | 36,000 |
| CWRU_12k_DE_Uni | NO | YES | YES | 11 | 60 | 120801-122917 | 1 | 7,313,758 |
| **CWRU_12k_DE_Multi** | NO | YES | YES | 9 | 52 | 120801-122917 | 3 | 19,013,055 |
| CWRU_12k_FE_Uni | NO | YES | YES | 9 | 45 | 120617-122269 | 1 | 5,458,059 |
| **CWRU_12k_FE_Multi** | NO | YES | YES | 9 | 45 | 120617-122269 | 3 | 16,374,177 |
| CWRU_48k_DE_Uni | NO | YES | YES | 9 | 52 | 63788-491446 | 1 | 21,416,809 |
| **CWRU_48k_DE_Multi** | NO | YES | YES | 9 | 52 | 63788-491446 | 2 | 42,833,618 |

Table 22. Dataset final characteristics after preprocessing

| Dataset Name | Varying Length | Scaling | (N) | (T) | (M) | # Instances |
|---|---|---|---|---|---|---|
| **BEARING_Uni** | N/A | Standardize | 2,560 | 8192 | 1 | 20,971,520 |
| **PHM2022_Multi** | Post padding to the max with 0 | N/A | 53,785 | 749 | 3 | 120,854,895 |
| PHM2022_PIN_Uni | Post padding to the max with 0 | N/A | 53,785 | 749 | 1 | 40,284,965 |
| PHM2022_PO_Uni | Post padding to the max with 0 | N/A | 53,785 | 749 | 1 | 40,284,965 |
| PHM2022_PDIN_Uni | Post padding to the max with 0 | N/A | 53,785 | 749 | 1 | 40,284,965 |
| **ETCHING_Multi** | Post padding to the max with mean | Standardize | 128 | 112 | 18 | 258,048 |
| **MFPT_48KHZ_Uni** | Frequency change & truncate end | N/A | 1,898 | 2000 | 1 | 3,796,000 |
| MFPT_96KHZ_Uni | Frequency change & truncate end | N/A | 1,898 | 4000 | 1 | 7,592,000 |
| PADERBORN_64KHZ_Uni | Truncate the end of the TS | N/A | 192,477 | 3000 | 1 | 577,431,000 |
| PADERBORN_4KHZ_Uni | Truncate the end of the TS | N/A | 185,520 | 200 | 1 | 37,104,000 |
| **PADERBORN_64KHZ_Multi** | Truncate the end of the TS | N/A | 192,477 | 3000 | 3 | 1,732,293,000 |
| PADERBORN_4KHZ_Multi | Truncate the end of the TS | N/A | 185,520 | 200 | 5 | 185,520,000 |
| Hydraulic_sys_10HZ_Multi | Scale all sensors to T=600 | Standardize | 2,205 | 600 | 17 | 22,491,000 |
| **Hydraulic_sys_100HZ_Multi** | Use sensors with length T=6000 | Standardize | 2,205 | 6000 | 17 | 92,610,000 |
| **Gas_sensors_home_activity** | Post padding to max with mean | Standardize | 99 | 15393 | 10 | 15,239,070 |
| **Control_charts** | N/A | Standardize | 600 | 60 | 1 | 36,000 |
| CWRU_12k_DE_Uni | Truncate end of the TS | N/A | 7200 | 1000 | 1 | 7,200,000 |
| **CWRU_12k_DE_Multi** | Truncate end of the TS | N/A | 6240 | 1000 | 3 | 18,720,000 |
| CWRU_12k_FE_Uni | Truncate end of the TS | N/A | 5400 | 1000 | 1 | 5,400,000 |
| **CWRU_12k_FE_Multi** | Truncate end of the TS | N/A | 5400 | 1000 | 3 | 16,200,000 |
| CWRU_48k_DE_Uni | Truncate end of the TS | N/A | 10681 | 2000 | 1 | 21,362,000 |
| **CWRU_48k_DE_Multi** | Truncate end of the TS | N/A | 10681 | 2000 | 2 | 42,724,000 |

There were some datasets that we were able to extract multiple datasets with different characteristics. For example in the case of the CWRU dataset, the data was collected from three sensors installed on the device namely drive end accelerometer data (DE), fan end accelerometer data (FE), and base accelerometer data (BA). The Data was collected at different sample rates (i.e., 12,000 samples/second and 48,000 samples/second) and the number of experiments in each of these conditions is different[6]. As a result, we extracted six different datasets with different characteristics from the raw data as illustrated in Table 22. We differentiated these datasets with different names for the sake of clarity. For instance "CWRU_12k_DE_uni" refers to the univariate dataset including only the data from the DE sensor gathered in the 12k sample rate.

The result is *22 preprocessed datasets* that are ready to be ingested by the selected TSC algorithms to evaluate their performance over multiple datasets. Based on Demsar et al., a number of datasets greater than ten are considered sufficient for this kind of performance evaluation[102]. Tables 21 and 22 show the details of dataset characteristics before and after the preprocessing step for reproducibility. The bolded items in Table 21 and Table 22 refer to a reduced set of eleven independent datasets that will be used as a scenario in the experiments.

[6] https://engineering.case.edu/bearingdatacenter/welcome

## 3.3. Experimental Setup

In this section, we illustrate the details of the experimental evaluation and the evaluation metrics.

### 3.3.1. Implementation of experimental evaluation

We implemented our experiments in two separate Python environments. For conventional ML algorithms, we used Python 3.8 with scikit-learn 1.2.2, sktime 0.18.0, tsfresh 0.20.0, tslearn 0.5.3.2, xgboost 1.7.5, mrsqm 0.0.1, pyts 0.12.0 as major ML python packages and for DL algorithms we used Python 3.9 with tensorflow 2.11.0 and torch 2.0.0. All these experiments have been done on an AMD Threadripper Pro 5975WX with 32 cores (64 threads), with 2x RTX A5000 (24 GB) GPUs, and 128 GB or 512GB of memory. The difference in memory available is due to an upgrade made necessary by some of the more demanding algorithm/data set combinations. We set a 48-hour runtime limit for each time running an algorithm on a dataset. This seems reasonable given that each of the 36 algorithms was run five times on each of the 22 datasets and running this number of experiments for 48 hours on a single CPU core would accumulate to approximately 21 years. In practice, we conducted experiments equivalent to approximately 2 years. These time limits apply to running the algorithm once on a single dataset. The algorithms were stopped manually after exceeding the runtime limit, and an accuracy of zero was recorded for them in calculations moving forward. Additionally, there were instances when the system ran out of memory. This decision is again based on the study's purpose to provide practical guidance to readers.

Our goal is to test the algorithms in their default parameter setting and how well they can generalize on manufacturing datasets with the minimum amount of parameter or hyperparameter tuning. So we did not carry out any parameter or hyperparameter tuning and we performed them on the default setting proposed by the referenced paper or the Python package authors. Unless explicitly stated in the original paper proposing the algorithm, we have not conducted any external tuning. More details for each algorithm can be found in the associated papers. While we acknowledge that this choice might impact some algorithms more than others, we believe it to be the best way to avoid additional bias and in accordance with our objective to provide insights to practitioners and academics alike.

To avoid categorizing algorithms into separate groups based on their capability to handle univariate or multivariate TSC, we adopted a different approach in our methodology. We treated the ability to handle multivariate time-series as an internal capability of the algorithm. In cases where an algorithm did not possess the multivariate capability, we used only the first dimension of the dataset as input for that specific algorithm. By doing so, we ensured a consistent experimental setup throughout our evaluation.

To be able to better evaluate the algorithms on unseen data, we have trained each algorithm five times on each dataset by using a five-fold cross-validation approach. Each fold uses a different 80/20 percent train/test split with a different random shuffled initialization, which enables us to take the mean accuracy and standard deviation over the five runs to reduce the bias due to the initial values and provide an analysis of uncertainty and variability. For DL algorithms, we ensured that all models converged during the training phase. We achieved this by choosing a high number of epochs for the given architecture and applying an early stopping callback if the model loss did not improve for 50 epochs.

## 3.3.2. Evaluation metrics

To evaluate the performance of different algorithms on multiple datasets, we followed the recommendations of Demsar[102], and we adopted the non-parametric *Friedman test* [102] to reject the null hypothesis. The null hypothesis being tested is that all classifiers perform the same and the observed differences are merely random. Then, the significance of differences between compared classifiers is measured using pairwise post-hoc analysis by a *Wilcoxon signed-rank* test with Holm correction ($\alpha$ = 0.05). A *critical difference (CD) diagram* is used to intuitively visualize the performance of these classifiers[103]. Since some of our datasets are imbalanced, we use the *weighted F1 score* as the measurement of accuracy.

Moreover, we followed the recommendations by Wang et al,.[53] and reported the *"Average Accuracy", "Number of Wins", "Average Rank"*, and the *"Mean Per-Class Error (MPCE)"* evaluation metrics based on classification error. Average accuracy is the average statistic of a given algorithm over all datasets. The number of wins indicates the number of times that a given algorithm outperformed all other algorithms, counting the ties for all winning algorithms. The average rank is defined to measure the algorithm's difference over multiple datasets. MPCE is a robust baseline criterion that calculates the mean error rates by considering the number of classes and can provide additional insights. The equation is as follows:

$$MPCE = \frac{1}{K}\sum_{k=1}^{k}\frac{e_k}{D_k}$$

Where K is the number of datasets, $D_k$ represents the number of classes in the dataset k, and $e_k$ represents the error rate on the k-th dataset. These two evaluation approaches are widely accepted among the TSC community.

Tables 23 and 24 summarize the parameters and configurations of the algorithm used to generate the results. We also documented the multivariate capability, which processing units were used for any of the algorithms, and the parallelization capability of each algorithm. While there may be other implementations of these algorithms with different characteristics, this table records the specific implementations used in this study. For the algorithms capable of running on GPU, we ran the algorithms on all available GPU cores, and for the algorithms running on CPU, we used 20 CPU cores for each algorithm. More details about each algorithm can be found in the referenced Python libraries.

Table 23. Conventional ML TSC algorithms parameters. "Multivariate capability" refers to whether the current implementation of the algorithm can process three-dimensional multivariate time-series data. "CPU/GPU" indicates the type of processing units employed for conducting our experiments. Lastly, "parallelization" denotes whether the implementation supports the execution of the algorithm across multiple cores, either on the CPU or GPU.

| Algorithm Name | Source | Algorithm parameters | Multivariate Capability | CPU/ GPU | Parallelization |
|---|---|---|---|---|---|
| LR | Sklearn | C=10 | NO | CPU | YES |
| NB | Sklearn | None | NO | CPU | NO |
| Ridge | Sklearn | None | No | CPU | NO |
| RF | Sklearn | n_estimators=500 | NO | CPU | YES |
| SVM | Sklearn | C=10, gamma='auto' | NO | CPU | NO |
| KNN-EUC | SKTIME[7] | distance = 'euclidean' | YES | CPU | YES |
| PF | SKTIME | n_estimators= 50, n_stump_evaluations=5 | NO | CPU | YES |
| KNN-TWE | SKTIME | distance = 'twe' | YES | CPU | NO |
| KNN-DTW-I | SKTIME | distance = 'dtw' | YES | CPU | NO |
| EE | SKTIME | distance_measures = "all",majority_vote=True | NO | CPU | YES |
| XGBoost | XGBOOST[8] | n_estimators=100, max_depth=6 | NO | CPU | YES |
| KNN-DDTW | SKTIME | distance = 'ddtw' | YES | CPU | NO |
| FBL | Based on [22] | tsfresh_features =EfficientFCParameters | YES | CPU | YES |
| BOSS-VS | PYTS[9] | n_bins=nb_classes, window_size=20 | YES | CPU | YES |
| DTWF | PYTS | dist="square", method = "sakoechiba", options={'window_size': 0.1}, SVM_kernel='poly | NO | CPU | YES |
| TDE | SKTIME | n_parameter_samples= 250, max_ensemble_size=50, randomly_selected_params = 50 | YES | CPU | YES |
| MrSQM | SKTIME | strat='RS', features_per_rep=500, selection_per_rep=2000 | NO | CPU | NO |
| LS | SKTIME | n_shapelets_per_size=0.1, min_shapelet_length=0.05, C=100 | NO | CPU | YES |
| STC | SKTIME | max_shapelets=1000, batch_size=100 | YES | CPU | YES |
| DrCIF | SKTIME | n_estimators=200, att_subsample_size= 10 | YES | CPU | YES |
| RISE | SKTIME | n_estimators = 200 | NO | CPU | YES |
| RotF | SKTIME | n_estimators = 50 | NO | CPU | YES |
| ARSENAL | SKTIME | num_kernels= 2000, n_estimators= 25, rocket_transform= "rocket" | YES | CPU | YES |
| ROCKET | SKTIME | n_kernel=1000 | YES | CPU | YES |
| HIVE-COTE 2 | SKTIME | default | YES | CPU | YES |

[7] https://github.com/sktime/sktime
[8] https://github.com/dmlc/xgboost
[9] https://pyts.readthedocs.io/en/stable/index.html

Table 24. ANN and DL TSC algorithms parameters

| Algorithm Name | Source | Algorithm parameters | Multivariate Capability | CPU/ GPU | Parallelization |
|---|---|---|---|---|---|
| MLP | dl-4-tsc[10] | Optimization= AdaDelta, Loss= Entropy, Epochs= 1000, Batch=16, Learning rate = 0.1 | YES | GPU | YES |
| FCN | dl-4-tsc | Optimization= Adam, Loss= Entropy, Epochs= 1000, Batch=16, Learning rate = 0.001 | YES | GPU | YES |
| Encoder | dl-4-tsc | Optimization= Adam, Loss= Entropy, Epochs= 100, Batch=12, Learning rate = 0.00001 | YES | GPU | YES |
| ResNet | dl-4-tsc | Optimization= Adam, Loss= Entropy, Epochs= 1000, Batch=32, Learning rate = 0.001 | YES | GPU | YES |
| InceptionTime | dl-4-tsc | Optimization= Adam, Loss= Entropy, Epochs= 1000, Batch=32, Learning rate = 0.001, kernel_size=41 | YES | GPU | YES |
| Stacked LSTM | Based on [69] | Optimization= RMSprop, Loss= Entropy, Epochs= 200, Batch=64, Learning rate = 0.001 | YES | GPU | YES |
| BI-LSTM | Based on [70] | Optimization= RMSprop, Loss= Entropy, Epochs= 500, Batch=64, Learning rate = 0.001 | YES | GPU | YES |
| TS-LSTM | Based on [65] | Optimization= Adam, Loss= Entropy, Epochs= 500, Batch=64, Learning rate = 0.001 | YES | GPU | YES |
| DA-NET | DANET[11] | Optimization= Adam, Loss= Entropy, Epochs= 100, Batch=16, Learning rate = 0.001 | YES | GPU | YES |
| MALSTM-FCN | MLSTM-FCN[12] | Optimization= Adam, Loss= Entropy, Epochs= 100, Batch=128, Learning rate = 0.001 | YES | GPU | YES |
| GASF-CNN | Based on [67] | Optimization Algorithm= Adam, Loss= Entropy, Epochs= 500, Batch=64, Learning rate = 0.001 | YES | GPU | YES |

# 4. Results & Discussion

In this section, we will present the results and analysis of our implemented experimental evaluation from different perspectives and offer additional insights. We were not obtaining results for all algorithms on all datasets within our defined constraints. Our objective in this evaluation was to assess the performance of classifiers based on the original authors' (or default) recommended configurations without any optimization. While it is possible that we could have tailored these algorithms' parameters to function more accurately on the challenging datasets, our intention was to avoid introducing bias into our results by doing so. Instead, we aim to determine which algorithms demonstrate better generalization capabilities across different problems and exhibit robust performance without dataset-specific optimizations.

[10] https://github.com/hfawaz/dl-4-tsc
[11] https://github.com/Sample-design-alt/DANet
[12] https://github.com/houshd/MLSTM-FCN

Table 25 shows the algorithms' accuracy on all datasets. Each number is the average statistic over five runs. Moreover, it contains the overall average accuracy (AVG ACC), number of wins (WIN), average rank (AVG Rank), and the mean per-class error (MPCE) metrics. The results ‘‘time” in the table denote that the corresponding algorithm failed to generate any result within defined time and resource constraints, and “OOM” indicates that the algorithm was stopped due to needing more than available 512GB memory. The dataset names have been abbreviated to be able to fit into the table. Table 25 numbers are the results of approximately two years of experiment runtime.

The ResNet, DrCIF, InceptionTime, and ARSENAL algorithms are showing the best overall performance in our experience. They were all able to achieve an average accuracy of higher than 96.6% on 22 datasets which is very impressive. The ResNet algorithm achieved the best results on WIN and AVG Rank metrics and DrCIF achieved the best results in AVG ACC and MPCE. The InceptionTime and ARSENAL algorithms are also showing overall competitive results in this experiment.

The DrCIF and ARSENAL algorithms belong to the conventional ML algorithms category proving that DL algorithms are not always the best solution and there are very powerful algorithms among ML algorithms as well. DrCIF employs interval-based feature extraction techniques and derives a set of features referred to as catch22 features, in addition to summary statistic features obtained from intervals within the time series. Subsequently, it leverages a DT ensemble classification technique for classification purposes. On the other hand, ARSENAL adopts an ensemble approach by employing multiple ROCKET algorithms for classification. Each ROCKET classifier utilizes a range of convolution kernels for feature extraction and relies on the ridge classifier for carrying out the classification process.

These results also show the robustness, efficiency, scalability, and power of convolution kernels in capturing temporal features in time-series data as three out of four best-performing algorithms are using these kernels for feature extraction.

The LSTM, BiLSTM, and TS-LSTM algorithms are another noteworthy group of algorithms in this experiment. These algorithms are based on RNN architectures and are showing comparatively very good overall performance which shows the effectiveness of RNN-based structures in capturing features in time-series data.

One particularly interesting outcome of this experiment is the underperformance of certain algorithms that are typically considered state-of-the-art in the TSC literature. Specifically, algorithms such as KNN-DTW, HIVE-COTE V2.0, and STC faced limitations while passing the experiment’s defined maximum runtime, resulting in poor overall performance. It is quite plausible that these algorithms could have demonstrated remarkable accuracy had we been able to run them continuously until obtaining conclusive outcomes. Notably, HIVE-COTE V2.0 impressively achieved 100% accuracy in two out of three instances when it managed to produce any results at all. These findings likely stem from the unique characteristics of the manufacturing datasets employed in this study, characterized by long time-series lengths and a substantial number of instances. We believe that this result holds important implications for both researchers and practitioners engaged in TSC tasks within smart manufacturing systems, offering valuable insights for their endeavors.

Table 25. Accuracy result of all 36 algorithms on 22 datasets. Non-highlighted results are obtained with 128GB of memory and the gray highlighted results are obtained with 512GB RAM. The "time" means the algorithms failed to produce results in the defined time constraints. And "OOM" means the system ran out of memory. All OOM was run on 512GB. The highest number in each row is indicated with bold font.

| Dataset Name | Ridge | LR | NB | RF | SVM | KNN-EU | PF | MLP | NN-DDTW | FBL | KNN-TWE | BOSSVS | LS | DrCIF | DTWF | RotF | XGBoost | STC | RISE |
|---|---|---|---|---|---|---|---|---|---|---|---|---|---|---|---|---|---|---|---|
| BEARING_U | 0.1788 | 0.1554 | 0.1486 | 0.4233 | 0.4167 | 0.4400 | time | 0.3279 | time | 0.9816 | time | 0.4739 | time | 0.9996 | time | 0.2078 | 0.3671 | time | **1.0000** |
| PHM22_M | 0.8152 | 0.9262 | 0.5158 | 0.9826 | 0.9866 | 0.9719 | OOM | 0.9972 | time | 0.5747 | time | 0.0301 | time | 0.9980 | time | 0.9044 | 0.9874 | time | 0.9819 |
| PHM22_PIN_U | 0.8774 | 0.9544 | 0.5118 | 0.9823 | 0.9647 | 0.9164 | OOM | 0.9875 | time | 0.7909 | time | 0.2949 | time | 0.9930 | time | 0.9112 | 0.9877 | time | 0.9889 |
| PHM22_PO_U | 0.8507 | 0.9141 | 0.4225 | 0.9731 | 0.8735 | 0.9264 | OOM | 0.8336 | time | 0.7605 | time | 0.1929 | time | 0.9917 | time | 0.8765 | 0.9801 | time | 0.9771 |
| PHM22_PDIN_U | 0.8145 | 0.9266 | 0.5163 | 0.9828 | 0.9863 | 0.9534 | OOM | 0.9883 | time | 0.7720 | time | 0.0302 | time | 0.9879 | time | 0.9040 | 0.9870 | time | 0.9809 |
| ETCHING_M | 0.6203 | 0.7081 | 0.5491 | 0.7703 | 0.7699 | 0.8041 | 0.7868 | 0.7743 | 0.8054 | **0.8913** | 0.8183 | 0.7563 | 0.7735 | 0.8706 | 0.7505 | 0.7223 | 0.7639 | 0.8795 | 0.7897 |
| MFPT_48_U | 0.4719 | 0.5685 | 0.7608 | 0.6411 | 0.7556 | 0.4032 | OOM | 0.6148 | 0.6623 | 0.9640 | 0.6345 | 0.6302 | 0.3478 | 0.9995 | 0.9984 | 0.5462 | 0.6173 | **1.0000** | **1.0000** |
| MFPT_96_U | 0.5229 | 0.5795 | 0.6776 | 0.6409 | 0.7459 | 0.4122 | OOM | 0.5909 | 0.6627 | 0.9774 | 0.6734 | 0.8164 | time | 0.9995 | 0.9989 | 0.5646 | 0.5711 | **1.0000** | **1.0000** |
| PADER_64_U | 0.4622 | 0.4875 | 0.2574 | 0.5354 | 0.6734 | 0.4218 | OOM | 0.8305 | time | 0.6230 | time | OOM | time | 0.9814 | time | 0.4307 | 0.6893 | time | 0.9660 |
| PADER_4_U | 0.4885 | 0.4978 | 0.3642 | 0.6703 | 0.5933 | 0.6501 | OOM | 0.8579 | time | 0.7788 | time | 0.3105 | time | 0.9251 | time | 0.5240 | 0.7231 | time | 0.9030 |
| PADER_64_M | 0.4601 | 0.4874 | 0.2580 | 0.5358 | 0.6726 | 0.4088 | OOM | 0.8829 | time | time | time | OOM | time | 0.9958 | time | 0.4312 | 0.6885 | time | 0.9648 |
| PADER_4_M | 0.4895 | 0.4976 | 0.3647 | 0.6697 | 0.5936 | 0.9086 | OOM | 0.9405 | time | 0.6144 | time | 0.3106 | time | 0.9955 | time | 0.5229 | 0.7233 | time | 0.9034 |
| Hydra_10_M | 0.8050 | 0.9420 | 0.1774 | 0.9723 | 0.6795 | 0.5587 | OOM | 0.7392 | 0.2185 | 0.7633 | 0.5414 | 0.2928 | 0.2462 | **0.9859** | 0.3291 | 0.9168 | 0.9660 | 0.9815 | 0.9805 |
| Hydra_100_M | 0.8680 | 0.9638 | 0.1774 | 0.9615 | 0.6766 | 0.5808 | OOM | 0.8228 | 0.6068 | 0.6908 | time | 0.5646 | time | 0.9896 | time | 0.9023 | 0.9641 | **0.9918** | 0.9822 |
| Gas_sensors | 0.5328 | 0.5790 | 0.4755 | 0.6267 | 0.6464 | 0.5555 | 0.5306 | 0.7427 | 0.6132 | 0.6712 | 0.4577 | 0.4770 | time | **0.8389** | 0.3435 | 0.6386 | 0.7096 | 0.8384 | 0.6796 |
| Control_charts | 0.7585 | 0.9414 | 0.9685 | 0.9783 | 0.980 | 0.9023 | 0.4695 | 0.9502 | 0.5894 | 0.9750 | 0.9816 | 0.9683 | 0.9784 | 0.9967 | 0.9917 | 0.8380 | 0.9231 | 0.9933 | 0.6647 |
| CWRU_12D_U | 0.2107 | 0.2375 | 0.5094 | 0.5443 | 0.6161 | 0.4323 | OOM | 0.7756 | 0.9639 | 0.9933 | 0.9313 | 0.3104 | time | 0.9999 | time | 0.3829 | 0.7418 | 0.9976 | 0.9997 |
| CWRU_12D_M | 0.2566 | 0.2488 | 0.5250 | 0.4894 | 0.5653 | 0.4835 | OOM | 0.8253 | 0.8873 | 0.9986 | 0.9219 | 0.4525 | time | **1.0000** | time | 0.3835 | 0.7075 | 0.9995 | 0.9995 |
| CWRU_12F_U | 0.3180 | 0.3131 | 0.2489 | 0.2664 | 0.3532 | 0.3322 | OOM | 0.7276 | 0.8569 | 0.9933 | 0.8638 | 0.3350 | time | 0.9998 | 0.5357 | 0.2641 | 0.5599 | 0.9981 | 0.9998 |
| CWRU_12F_M | 0.3696 | 0.3618 | 0.3383 | 0.3014 | 0.3266 | 0.5437 | OOM | 0.8567 | 0.9014 | 0.9978 | 0.8863 | 0.5251 | 0.1903 | **1.0000** | 0.6931 | 0.2709 | 0.4666 | 0.9998 | 0.9994 |
| CWRU_48D_U | 0.2764 | 0.3122 | 0.2541 | 0.6875 | 0.6562 | 0.5416 | OOM | 0.8908 | time | 0.8974 | time | 0.4001 | time | 0.9987 | time | 0.3765 | 0.8097 | 0.9981 | 0.9984 |
| CWRU_48D_M | 0.2785 | 0.3112 | 0.2536 | 0.6881 | 0.6474 | 0.5307 | OOM | 0.9147 | time | 0.9578 | time | 0.4000 | time | 0.9997 | time | 0.3831 | 0.8141 | 0.8862 | 0.9980 |
| AVG ACC | 0.533 | 0.587 | 0.422 | 0.697 | 0.690 | 0.622 | 0.081 | 0.812 | 0.353 | 0.803 | 0.350 | 0.390 | 0.115 | **0.979** | 0.256 | 0.586 | 0.761 | 0.571 | 0.9440 |
| WIN | 0 | 0 | 0 | 0 | 0 | 0 | 0 | 0 | 0 | 1 | 0 | 0 | 0 | 4 | 0 | 0 | 0 | 3 | 3 |
| AVG Rank | 25.14 | 22.91 | 26.14 | 19.45 | 20.23 | 22.59 | 31.93 | 17.77 | 26.32 | 16.36 | 26.16 | 25.73 | 31.25 | 5.45 | 28.09 | 23.68 | 18.07 | 16.34 | 10.11 |
| MPCE | 0.096 | 0.085 | 0.119 | 0.065 | 0.066 | 0.083 | 0.180 | 0.043 | 0.130 | 0.050 | 0.133 | 0.120 | 0.175 | **0.007** | 0.139 | 0.086 | 0.054 | 0.088 | 0.017 |

| Dataset Name | ARSNL | KNNDTWI | TDE | EE | H-COTE2 | MrSQM | ROCKET | FCN | ENCODER | ResNet | InceptionTime | LSTM | Bi-LSTM | TS-LSTM | DA-NET | MALSTM-FCN | GASF-CNN |
|---|---|---|---|---|---|---|---|---|---|---|---|---|---|---|---|---|---|
| BEARING_U | **1.0000** | time | time | time | time | **1.0000** | **1.0000** | 0.9996 | 0.2877 | 0.9996 | 0.9996 | 0.9980 | 0.8312 | 0.9934 | 0.6917 | 0.5095 | 0.9011 |
| PHM22_M | 0.9985 | time | time | time | time | 0.9830 | 0.9979 | 0.9983 | 0.9978 | 0.9986 | **0.9986** | 0.9941 | 0.9979 | 0.9982 | 0.9817 | 0.9974 | 0.9976 |
| PHM22_PIN_U | 0.9966 | time | time | time | time | 0.9957 | 0.9933 | 0.9980 | 0.9953 | 0.9983 | 0.9775 | **0.9997** | 0.9955 | 0.9972 | 0.9838 | 0.9957 | 0.9936 |
| PHM22_PO_U | 0.9917 | time | time | time | time | 0.9908 | 0.9801 | 0.9952 | 0.9889 | **0.9974** | 0.9895 | 0.9964 | 0.9919 | 0.9852 | 0.9013 | 0.9824 | 0.9931 |
| PHM22_PDIN_U | 0.9969 | time | time | time | time | 0.9807 | 0.9916 | 0.9968 | 0.9972 | **0.9981** | 0.9956 | 0.9889 | 0.9928 | 0.9966 | 0.9146 | 0.9947 | 0.9954 |
| ETCHING_M | 0.8623 | 0.7505 | 0.7901 | 0.7669 | 0.8064 | 0.7865 | 0.8049 | 0.8766 | 0.8036 | 0.8293 | 0.8637 | 0.8013 | 0.8165 | 0.8303 | 0.7984 | 0.7978 | 0.8205 |
| MFPT_48_U | **1.0000** | 0.6735 | time | time | **1.0000** | 0.9989 | 0.9979 | **1.0000** | 0.9989 | **1.0000** | **1.0000** | 0.9984 | 0.9544 | 0.9963 | 0.7557 | 0.6862 | 0.9231 |
| MFPT_96_U | **1.0000** | 0.6703 | time | time | time | 0.9995 | 0.9968 | **1.0000** | 0.8624 | **1.0000** | **1.0000** | 0.9995 | 0.8871 | 0.9995 | 0.7526 | 0.6762 | 0.9244 |
| PADER_64_U | 0.9443 | time | time | time | time | 0.8387 | 0.8937 | 0.9258 | 0.9730 | **0.9963** | 0.9950 | 0.9719 | 0.9864 | 0.9865 | 0.6138 | 0.8373 | 0.7063 |
| PADER_4_U | 0.8376 | time | time | time | time | 0.7261 | 0.7893 | 0.9300 | 0.9321 | 0.9622 | **0.9827** | **0.9794** | 0.9622 | **0.9772** | 0.6518 | 0.8776 | 0.7244 |
| PADER_64_M | 0.9524 | time | time | time | time | 0.8609 | 0.8900 | 0.9969 | 0.9623 | **0.9997** | 0.9895 | 0.9763 | 0.9863 | 0.9934 | 0.6265 | 0.9545 | OOM |
| PADER_4_M | 0.9649 | time | time | time | time | 0.7406 | 0.9077 | 0.9996 | 0.9994 | **0.9999** | 0.9998 | 0.9997 | 0.9989 | 0.9997 | 0.5912 | 0.9992 | OOM |
| Hydra_10_M | 0.9230 | 0.5337 | 0.9787 | **0.9842** | time | 0.8806 | 0.7695 | 0.8693 | 0.7566 | 0.9599 | 0.9280 | 0.4782 | 0.7386 | 0.8358 | 0.7007 | 0.7087 | 0.8895 |
| Hydra_100_M | 0.9575 | time | time | time | time | 0.8093 | 0.8371 | 0.9859 | 0.8234 | 0.9855 | 0.9873 | 0.4453 | 0.7717 | 0.9202 | 0.9214 | 0.8488 | 0.7524 |
| Gas_sensors | 0.8202 | 0.5541 | 0.6837 | time | **1.0000** | 0.6714 | 0.5784 | 0.5586 | 0.5721 | 0.6005 | 0.5481 | 0.4804 | 0.5677 | 0.6920 | 0.6731 | 0.5233 | 0.5412 |
| Control_charts | **1.0000** | 0.9950 | 0.9917 | 0.9799 | time | 0.8591 | 0.9967 | 0.9933 | 0.9934 | 0.9933 | **0.9983** | 0.9866 | 0.9933 | 0.9767 | 0.9717 | 0.9599 | 0.6611 |
| CWRU_12D_U | **1.0000** | 0.9587 | time | time | time | 0.9918 | **1.0000** | 0.9999 | 0.9842 | **1.0000** | **1.0000** | 0.9960 | 0.9948 | 0.9948 | 0.8827 | 0.8682 | 0.7515 |
| CWRU_12D_M | **1.0000** | 0.9378 | time | time | time | 0.9893 | **1.0000** | **1.0000** | 0.9915 | **1.0000** | **1.0000** | 0.9961 | 0.9915 | 0.9950 | 0.8851 | 0.8800 | 0.8683 |
| CWRU_12F_U | 0.9996 | 0.8592 | time | time | time | 0.9957 | 0.9993 | 0.9996 | 0.9861 | **1.0000** | **1.0000** | 0.9946 | 0.9920 | 0.9900 | 0.7678 | 0.7747 | 0.5329 |
| CWRU_12F_M | **1.0000** | 0.9317 | time | time | time | 0.9753 | 0.9998 | 0.9996 | 0.9959 | 0.9998 | **1.0000** | 0.9993 | 0.9980 | 0.9989 | 0.9612 | 0.9156 | 0.7536 |
| CWRU_48D_U | 0.9984 | time | time | time | time | 0.9786 | 0.9921 | 0.9556 | 0.9634 | 0.9997 | **0.9998** | 0.9985 | 0.9870 | 0.9941 | 0.8123 | 0.9148 | 0.8047 |
| CWRU_48D_M | **1.0000** | time | time | time | time | 0.9800 | **1.0000** | 0.9995 | 0.9952 | **0.9999** | 0.9998 | 0.9914 | 0.9010 | 0.9624 | 0.9562 | 0.9352 | 0.9051 |
| AVG ACC | 0.966 | 0.357 | 0.157 | 0.124 | 0.128 | 0.911 | 0.928 | 0.958 | 0.903 | 0.969 | 0.966 | 0.912 | 0.924 | 0.960 | 0.809 | 0.847 | 0.747 |
| WIN | 8 | 0 | 0 | 0 | 2 | 1 | 4 | 3 | 0 | **10** | 9 | 2 | 0 | 0 | 0 | 0 | 0 |
| AVG Rank | 5.55 | 26.09 | 28.48 | 30.18 | 28.86 | 13.27 | 9.89 | 6.59 | 11.68 | **4.16** | 5.34 | 10.02 | 11.43 | 8.89 | 17.50 | 15.70 | 17.64 |
| MPCE | 0.012 | 0.132 | 0.162 | 0.173 | 0.168 | 0.028 | 0.023 | 0.014 | 0.024 | 0.011 | 0.012 | 0.026 | 0.021 | 0.013 | 0.051 | 0.038 | 0.066 |

## 4.1. Benchmark Algorithms

Algorithms specifically designed for TSC tasks should provide improvements in terms of accuracy metrics compared to existing benchmark algorithms. Using a benchmark classifier that treats each series as a vector, without considering the potential autocorrelation, is the apparent starting point for TSC tasks. However, TSC problems have certain characteristics that make them challenging, such as long series with many redundant or correlated attributes, variable lengths, seasonality, trend, and non-stationarity and noises. Since standard classifiers may struggle with these characteristics, there have been efforts to design classifiers that can compensate for them. However, not all TSC problems will have these characteristics, and benchmarking against standard classifiers can provide insights into the datasets' characteristics. Table 26 summarizes the comparison of seven benchmark algorithms on 22 datasets and provides four metrics to fully evaluate different approaches

Table 26. Performance comparison of seven benchmark algorithms on 22 datasets.

| Metric | Algorithms | | | | | | |
|---|---|---|---|---|---|---|---|
| | LR | NB | RF | SVM | KNN-EUC | Ridge | MLP |
| AVG ACC | 0.587 | 0.422 | 0.697 | 0.690 | 0.622 | 0.533 | **0.813** |
| WIN | 0 | 1 | 3 | 2 | 2 | 0 | **14** |
| AVG Rank | 4.55 | 5.91 | 2.86 | 2.91 | 4.27 | 5.41 | **2.09** |
| MPCE | 0.085 | 0.119 | 0.065 | 0.066 | 0.083 | 0.096 | **0.043** |

Figure 11 shows the critical difference diagram for seven benchmark algorithms listed in Table 20. The cliques are formed using a pairwise Wilcoxon test. The existence of a clique between a pair of algorithms means that they are not significantly different from each other over tested datasets.

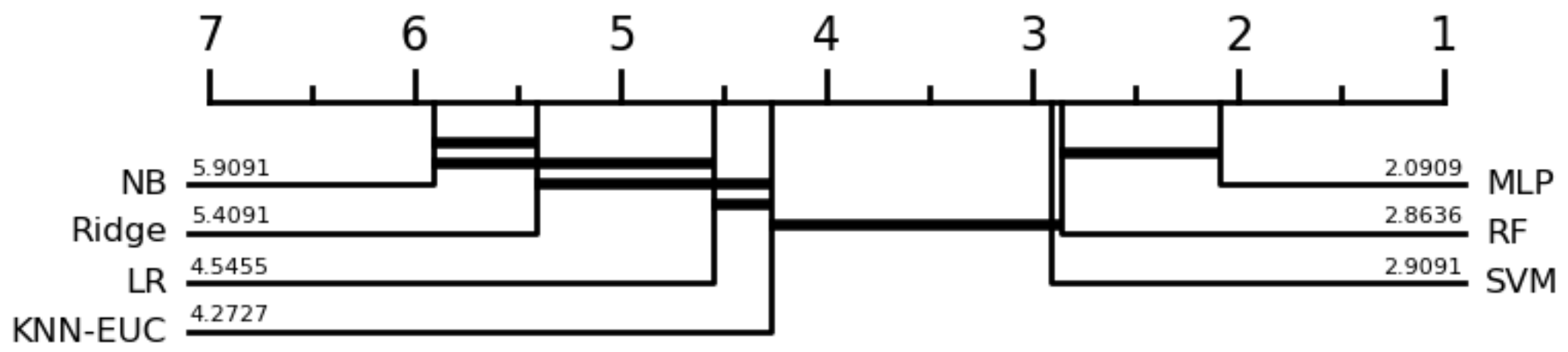

Figure 11: Critical difference diagrams for seven benchmark classifiers on the 22 datasets

Table 26 and Figure 11 consistently demonstrate that MLP outperforms the other seven benchmark algorithms, with RF coming in as the second-best performer. These results are calculated based on the raw accuracy measures provided in Table 25 and will not be repeated here. This outcome aligns with our expectations, given MLP's prowess in addressing complex and nonlinear problems, irrespective of any time-related factors.

## 4.2. Conventional TSC Algorithms

Due to the fundamental differences in structure and learning schema between conventional ML and DL algorithms, we conducted separate comparisons for each group to identify the best algorithms for manufacturing problems. In this group, we included 19 algorithms, of which all except XGBoost were explicitly designed to solve TSC problems. Due to the competitive performance of XGBoost in TSC classification tasks[104], it was included in this group alongside other algorithms. Additionally, we included RF as the top-performing non-DL algorithm based on the benchmark comparison. Table 27 summarizes the comparison results.

Table 27. Performance comparison of 20 Conventional algorithms on 22 datasets.

| Metric | Algorithms | | | | | | | | | |
|---|---|---|---|---|---|---|---|---|---|---|
| | RF | PF | KNN-DDTW | FBL | KNN-TWE | BOSSVS | LS | DrCIF | DTWF | RotF |
| AVG ACC | 0.697 | 0.081 | 0.353 | 0.803 | 0.350 | 0.390 | 0.115 | **0.979** | 0.256 | 0.586 |
| WIN | 0 | 0 | 0 | 1 | 0 | 0 | 0 | 11 | 0 | 0 |
| AVG Rank | 9.32 | 16.34 | 13.45 | 7.98 | 13.16 | 12.07 | 15.93 | **2.32** | 14.34 | 11.32 |
| MPCE | 0.065 | 0.180 | 0.130 | 0.050 | 0.133 | 0.120 | 0.175 | **0.007** | 0.139 | 0.086 |
| | XGBoost | STC | RISE | ARSENAL | KNN-DTW-I | TDE | EE | HIVE-COTE2 | MrSQM | ROCKET |
| AVG ACC | 0.761 | 0.571 | 0.944 | 0.966 | 0.357 | 0.157 | 0.124 | 0.128 | 0.911 | 0.928 |
| WIN | 0 | 3 | 4 | **12** | 0 | 0 | 0 | 2 | 1 | 4 |
| AVG Rank | 8.61 | 8.43 | 4.70 | 2.61 | 13.27 | 14.61 | 15.50 | 14.75 | 6.50 | 4.77 |
| MPCE | 0.054 | 0.088 | 0.017 | 0.012 | 0.132 | 0.162 | 0.173 | 0.168 | 0.028 | 0.023 |

Figure 12 presents the critical difference diagram for this analysis. We removed six worst-performer algorithms (PF, LS, DTWF, TDE, EE, and HIVE-COTE 2 ) from this figure to be able to have a more discernable figure. The difference between average ranks in Table 27 and Figure 12 is due to this decision aimed at making the results more digestible for the readers.

The DrCIF algorithm has demonstrated the best results among 20 conventional ML algorithms. It achieved the highest ranking in three out of four metrics in Table 27, as well as the best results in Figure 12. ARSENAL is another highly powerful algorithm. In Figure 12, it ranks as the second-best algorithm and is not significantly inferior to DrCIF. It also achieved the highest number of wins among all algorithms. These results underscore the effectiveness of both interval-based and kernel-based FE techniques when combined with ensemble learning classification techniques in TSC tasks.

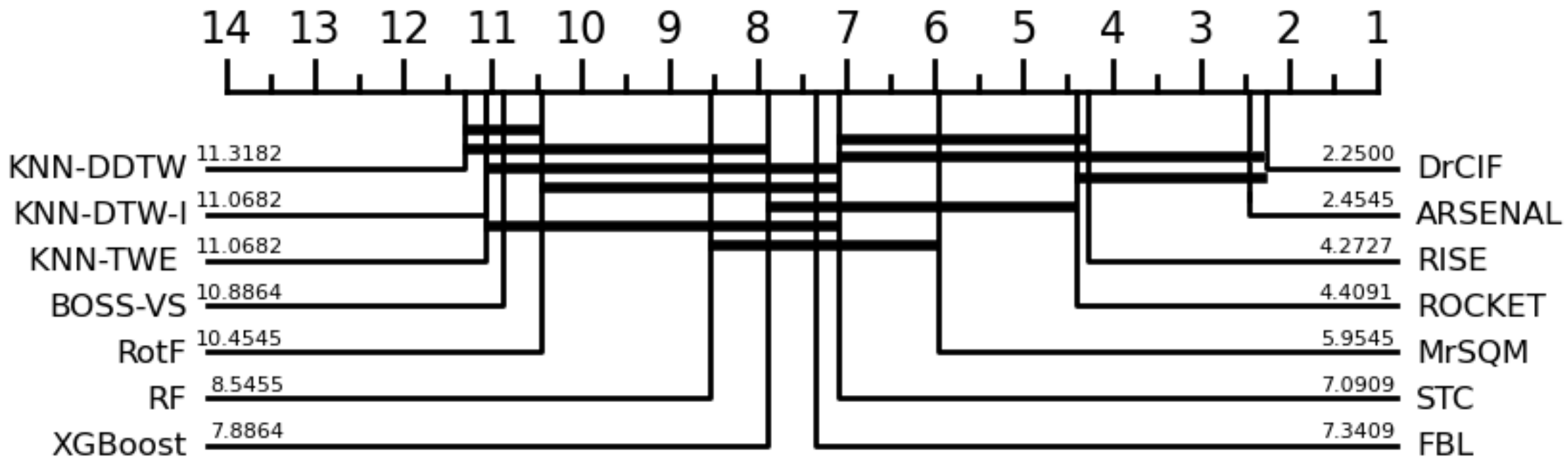

Figure 12: Critical difference diagrams for 20 conventional algorithms on the 22 datasets

## 4.3. ANN & DL Algorithms

Many reasons can be mentioned explaining the popularity of DL algorithms in recent years. DL algorithms excel at automatically learning and extracting features from the data, reducing the need for manual FE and extensive domain expertise. Moreover, their scalability and flexibility make them practical for handling large datasets and high-dimensional with long time-series data. Additionally, the utilization of advanced hardware and computational resources, such as GPUs has further facilitated the widespread adoption of DL algorithms across various domains. In this part, we assessed these capabilities to see how DL algorithms can approximate time-series problems in the manufacturing domain and find the best performers amongst them. We compared ten algorithms with different architectures, and MLP, as the benchmark in this group, was included in this group for comparison. Table 28 summarizes the comparison results. Except for MLP, DA-NET, GASF-CNN, and MASLSTM, all other tested DL algorithms are showing competitive results and AVG ACC and MPCE metrics. ResNet is superior in all four defined metrics and based on Table 28 results, and it can be considered the best ANN & DL algorithm for TSC tasks in the manufacturing domain.

Table 28. Performance comparison of eleven ANN & DL algorithms on 22 datasets.

| Metric | Algorithms | | | | | | | | | | |
|---|---|---|---|---|---|---|---|---|---|---|---|
| | MLP | FCN | Encoder | ResNet | Inception Time | LSTM | BiLSTM | TSLSTM | DA-NET | MALSTM-FCN | GASF-CNN |
| AVG ACC | 0.812 | 0.958 | 0.903 | **0.969** | 0.966 | 0.912 | 0.924 | 0.960 | 0.809 | 0.847 | 0.747 |
| WIN | 0 | 5 | 0 | **14** | 12 | 1 | 0 | 1 | 0 | 0 | 0 |
| AVG Rank | 9.23 | 3.55 | 6.07 | **2.07** | 2.77 | 5.64 | 6.30 | 4.61 | 8.82 | 8.36 | 8.59 |
| MPCE | 0.043 | 0.014 | 0.024 | **0.011** | 0.012 | 0.026 | 0.021 | 0.013 | 0.051 | 0.038 | 0.066 |

Figure 13 presents the critical difference diagram for this analysis. The results from this analysis agree with Table 28 and show that ResNet is superior among all eleven ANN & DL algorithms. InceptionTime, FCN, and TS-LSTM are in the next places respectively although they are not significantly different from ResNet based on the pairwise Wilcoxon test. This shows that they are also very powerful algorithms.

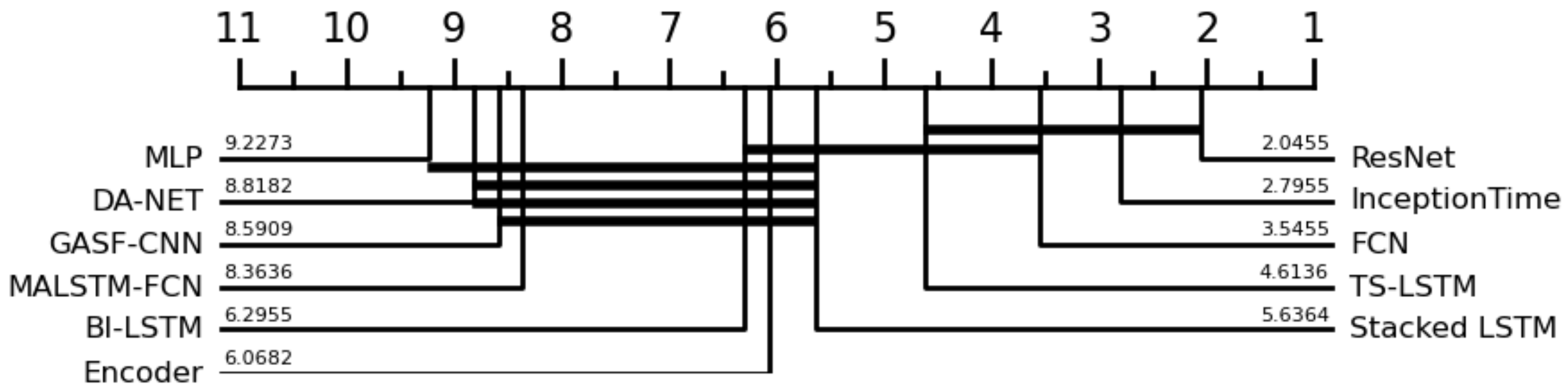

Figure 13: Critical difference diagrams for eleven DL algorithms on the 22 datasets

## 4.4. Results for Univariate Time-series (UTSC) and Multivariate Time-series (MTSC)

As mentioned earlier (refer to Figure 3), the TSC algorithms are viewed as comprehensive modules that receive raw time-series data with the shape *N*T*M* on one end and predict corresponding labels on the other end. We assume that the ability to handle multivariate time-series data and extract discriminative features from different dimensions of a given dataset is an internal capability of the algorithm. However, there may be situations where our focus is specifically on working with univariate time-series data, without requiring additional dimensions. It's worth noting that algorithms designed to handle multivariate data can naturally accept univariate data as well. The list of univariate and multivariate datasets can be found in Table 21.

To manage the size of the experiment effectively, we divided the datasets into two distinct groups. The first group of experiments focuses on testing the performance of the ten best-performing algorithms on twelve univariate datasets. Table 29 summarizes the comparison results. All compared algorithms show an average accuracy of more than 95% proving that all of them are capable of generating competitive results in UTSC tasks. The ResNet algorithm however shows superior performance in all four metrics. The InceptionTime is the second-best algorithm with tied results in two out of four metrics.

Table 29. Performance comparison of top 10 algorithms on 12 univariate datasets.

| Metric | Algorithms | | | | | | | | | |
|---|---|---|---|---|---|---|---|---|---|---|
| | ResNet | Inception Time | FCN | TSLSTM | LSTM | DrCIF | ARSENAL | RISE | ROCKET | BiLSTM |
| AVG ACC | **0.995** | **0.995** | 0.983 | 0.991 | 0.992 | 0.989 | 0.980 | 0.957 | 0.969 | 0.964 |
| WIN | **7** | 6 | 2 | 0 | 2 | 0 | 5 | 3 | 2 | 0 |
| AVG Rank | **2.75** | 3.62 | 5.21 | 6.87 | 5.75 | 5.71 | 4.33 | 6.50 | 6.92 | 7.33 |
| MPCE | **0.001** | **0.001** | 0.005 | 0.002 | 0.002 | 0.003 | 0.006 | 0.019 | 0.009 | 0.008 |

Figure 14 presents a critical difference diagram for this analysis, which indicates that there are no significant differences between any pairs of algorithms according to the pairwise Wilcoxon test. This suggests that all of these algorithms are capable of delivering satisfactory accuracy in UTSC tasks.

However, when considering the four metrics provided in Table 29, it becomes evident that the ResNet algorithm performs slightly better than the others, followed by InceptionTime, ARSENAL, and the FCN algorithm, in that order. Additionally, taking into account the runtime and computational expenses of these algorithms can serve as an additional factor for decision-making when distinguishing among equally competent algorithms. The results of this evaluation are elaborated upon in section 4.6.

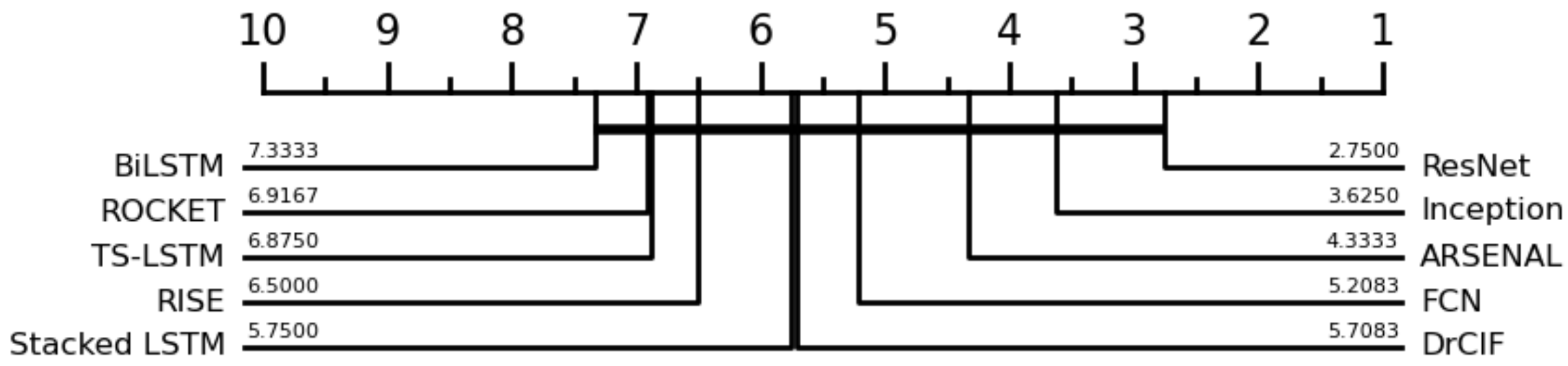

Figure 14: Critical difference diagrams for top 10 algorithms on 12 univariate (UTSC) datasets

The second group of experiments aims to evaluate the performance of the top ten best-performing algorithms on ten multivariate datasets. Table 30 summarizes the comparison results. Seven algorithms show an average accuracy of higher than 92% in this comparison. This shows that all these algorithms are competitive for MTSC tasks. However, DrCIF algorithms were able to outperform all other algorithms in all four metrics. ResNet, InceptionTime, and ARSENAL algorithms are in the next places respectively.

Table 30. Performance comparison of top 10 algorithms on 10 multivariate datasets.

| Metric | Algorithms | | | | | | | | | |
|---|---|---|---|---|---|---|---|---|---|---|
| | ResNet | Inception Time | FCN | TSLSTM | BiLSTM | DrCIF | ARSENAL | RISE | ROCKET | Encoder |
| AVG ACC | 0.937 | 0.931 | 0.928 | 0.923 | 0.877 | **0.967** | 0.948 | 0.928 | 0.879 | 0.890 |
| WIN | 4 | 3 | 2 | 0 | 0 | **5** | 3 | 0 | 2 | 0 |
| AVG Rank | 3.25 | 3.70 | 4.45 | 5.90 | 8.30 | **3.15** | 4.40 | 6.90 | 6.60 | 8.35 |
| MPCE | 0.023 | 0.024 | 0.025 | 0.026 | 0.038 | **0.013** | 0.019 | 0.027 | 0.040 | 0.036 |

Figure 15 displays the critical difference diagram for this analysis. While the differences between the compared algorithms are not significant enough to reject the null hypothesis in the pairwise Wilcoxon test, an examination of the metrics in Table 30 reveals that DrCIF slightly outperforms the other algorithms, with ResNet, InceptionTime, and ARSENAL algorithms following closely. Once more, the analysis of runtime and computational expenses in section 4.6 can provide additional insights for decision-making purposes.

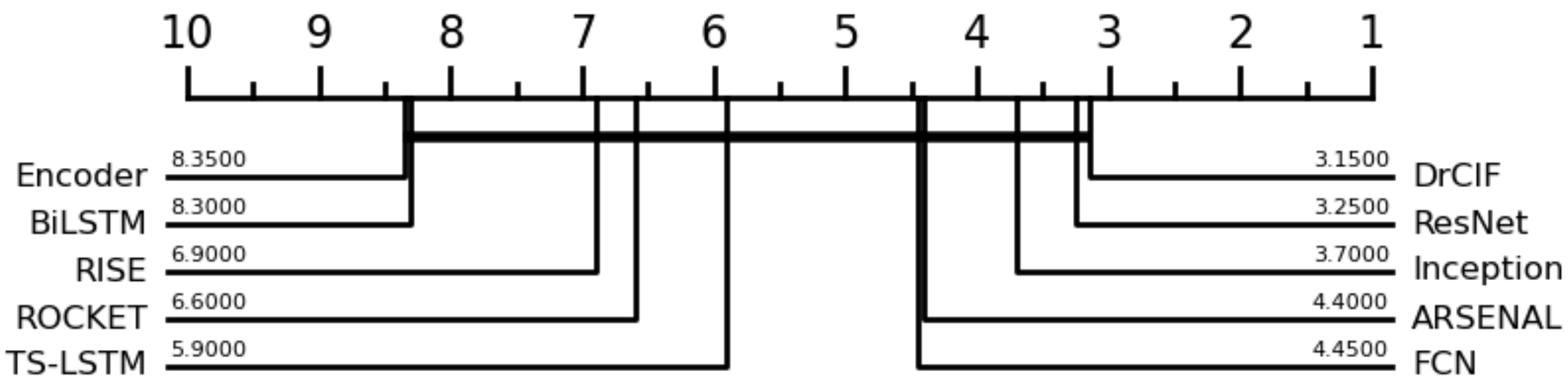

Figure 15: Critical difference diagrams for top 10 algorithms on 10 multivariate (MTSC) datasets

## 4.5. Results on reduced datasets

In this section, we conducted the experiment using a reduced set of eleven datasets. As part of the methodology, we augmented some of the datasets to create a more diverse dataset repository to reach the goal of covering a wider range of problems. For instance, the PHM2022 dataset was initially a multivariate dataset with three dimensions derived from three different sensors. We split each sensor's data into a univariate dataset and augmented it, resulting in four distinct datasets. In this analysis, we removed these augmented datasets to ensure a more distinct dataset collection and to test if the augmentation process introduced any biases. The relevant datasets are highlighted in bold font in Table 21.

To effectively manage the scale of the experiment, we tested the top ten best-performing algorithms in terms of AVG ACC metric on the mentioned eleven datasets. Table 31 provides a summary of the results. The results indicate that all these algorithms are performing very well on the reduced datasets. The DrCIF algorithm is superior in two metrics and the ARSENAL algorithm is superior in two metrics.

Table 31. Performance comparison of 10 best-performing algorithms on 11 datasets.

| Metric | Algorithms | | | | | | | | | |
|---|---|---|---|---|---|---|---|---|---|---|
| | DrCIF | RISE | ARSENAL | ROCKET | FCN | ResNet | Inception Time | BiLSTM | TSLSTM | MrSQM |
| AVG ACC | **0.972** | 0.915 | 0.963 | 0.918 | 0.946 | 0.946 | 0.944 | 0.892 | 0.942 | 0.901 |
| WIN | 4 | 2 | **6** | 3 | 3 | 4 | 4 | 0 | 0 | 1 |
| AVG Rank | 3.59 | 6.50 | **3.32** | 5.73 | 4.54 | 4.09 | 3.86 | 8.41 | 6.82 | 8.14 |
| MPCE | **0.011** | 0.026 | 0.014 | 0.029 | 0.020 | 0.020 | 0.021 | 0.032 | 0.020 | 0.031 |

Figure 16 presents the critical difference diagram for this analysis. Although the Wilcoxon test failed to reject the null hypothesis on most algorithm pairs, the ARSENAL and DrCIF algorithms are showing marginally better performance among others.

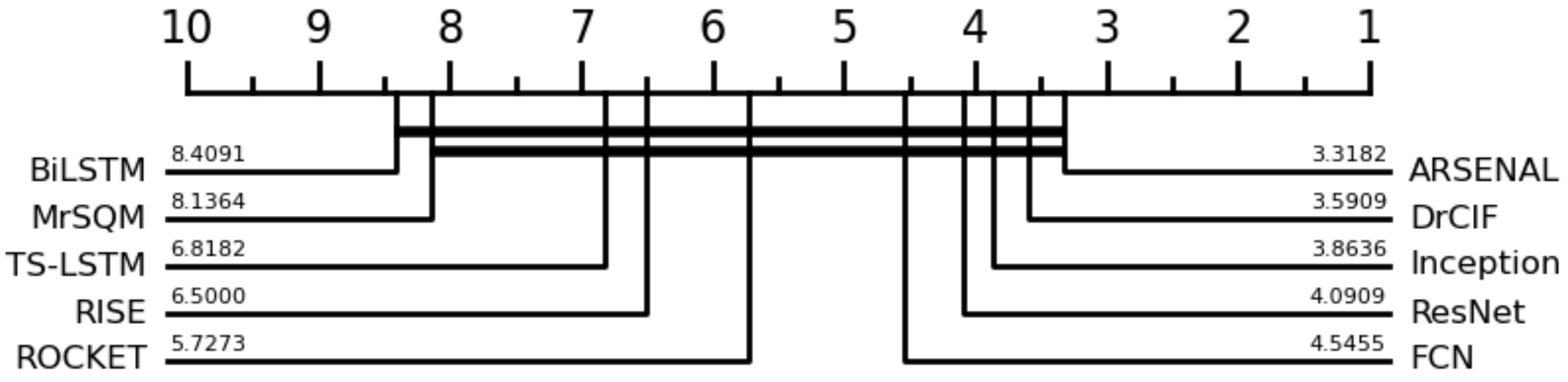

Figure 16: Critical difference diagrams for 10 algorithms on the 11 datasets

Figure 17 presents boxplots of 19 algorithms for all accuracy measures. These 19 algorithms were selected after removing those that failed to produce results for all datasets, as well as low-performing algorithms like LR, NB, and GASF-CNN, to enhance the clarity of the plot. These boxplots offer a visual summary of the distribution, skewness, and presence of outliers in accuracy measurements, aiding in the assessment of the reliability and robustness of different algorithms in comparison. For instance, ARSENAL and DrCIF exhibit consistent results with few outliers, while algorithms such as Bi-LSTM and MrSQM display larger boxes, indicating less consistency.

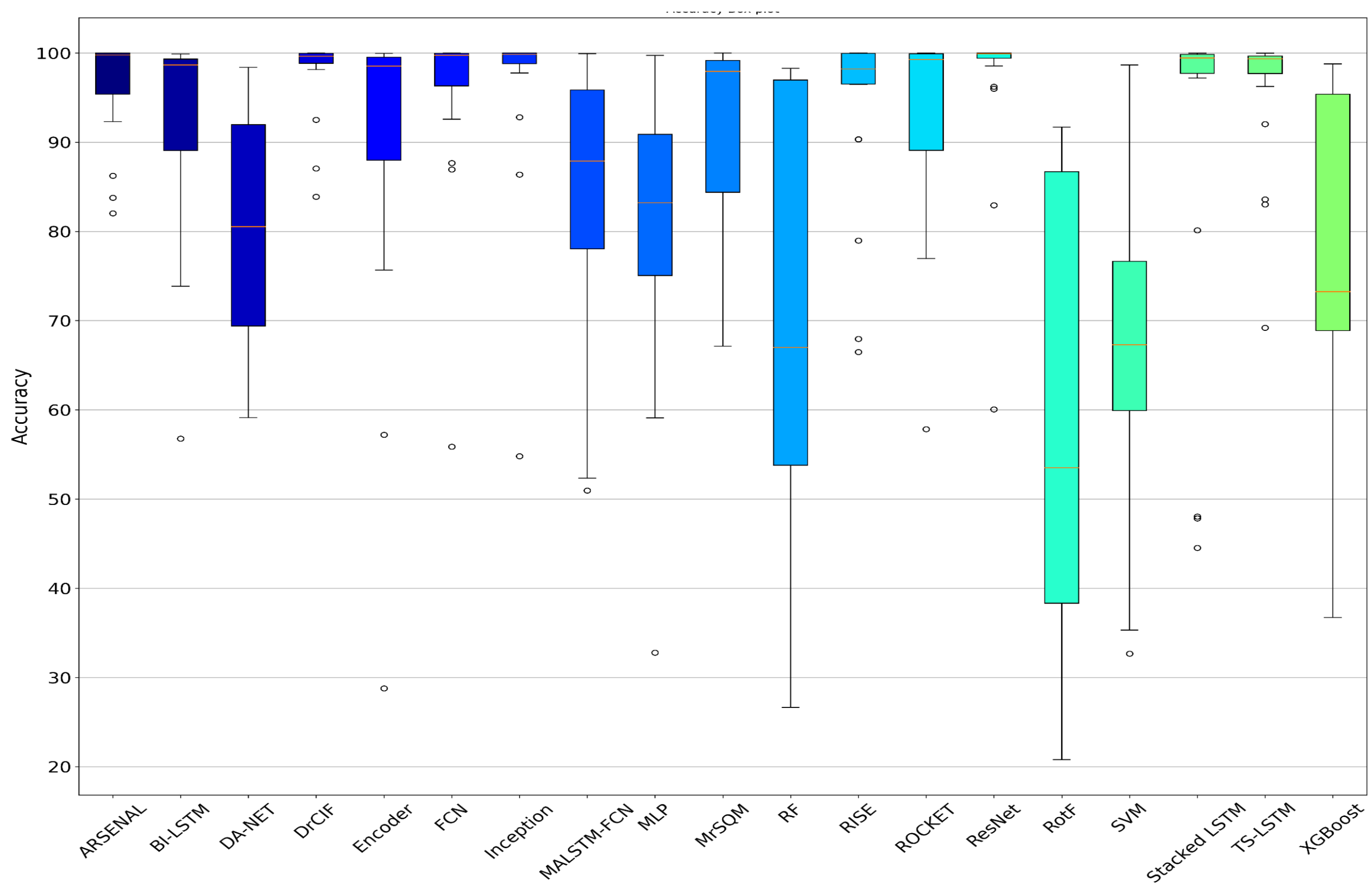

Figure 17: Accuracy Box plots for 19 algorithms on 22 datasets.

## 4.6. Runtime and Computational Expense Evaluation

As was shown in previous sections, there might be situations where we need extra evaluation metrics in addition to accuracy, to be able to make better decisions. Runtime can serve as such a metric and it can be directly related to the algorithms' computational complexity. While accuracy reflects an algorithm's ability to correctly classify data, runtime considerations can offer a different dimension of evaluation.

In real-world applications, especially those that require time-sensitive decisions or when we have limited available computation resources, the computational efficiency of an algorithm can be just as important as its accuracy. Faster algorithms with shorter runtimes are more practical for real-time and high-throughput systems, where quick decisions are imperative. They can operate efficiently on standard CPU configurations, removing the need for high-performance computers with advanced GPUs. Additionally, runtime assessments help identify trade-offs between computational complexity and accuracy, allowing practitioners to choose the most suitable algorithms based on their specific application requirements. Therefore, considering both accuracy and runtime in TSC tasks provides a more comprehensive perspective, enabling the selection of the right balance between classification performance and computational efficiency.

It is difficult to compare runtimes and computational expenses of different algorithms for several reasons such as differences in Python package software and available hardware resources differences. Moreover, some algorithms had been run on CPU while some others ran on GPU. We also ran several algorithms in parallel on different CPU and GPU cores and we do not know the effect of doing so on the final runtime (which was considered out of scope for this study). Although computational complexity assessments have been conducted by researchers over the years[28,31,105], our approach took a more practical route to gain insight into relative algorithm performance by recording runtime data for all experiments. Figure 18 provides the boxplot summaries of 19 algorithm runtime information and Figure 19 plots the average accuracy against runtime.

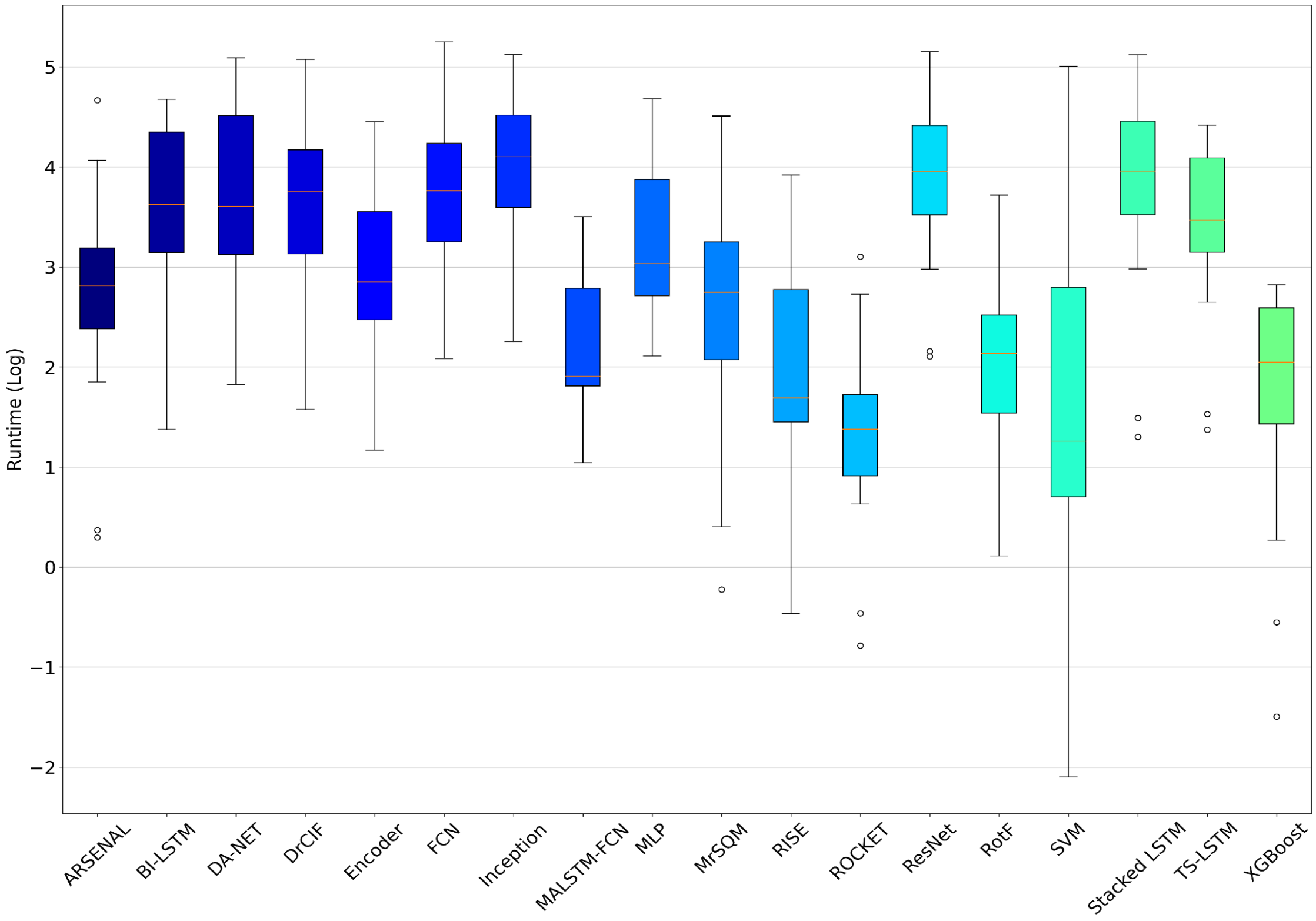

Figure 18: Runtime Box plots for 19 algorithms on 22 datasets. The vertical axis is the logarithmic transformation of runtime in seconds and the horizontal axis is the name of different algorithms.

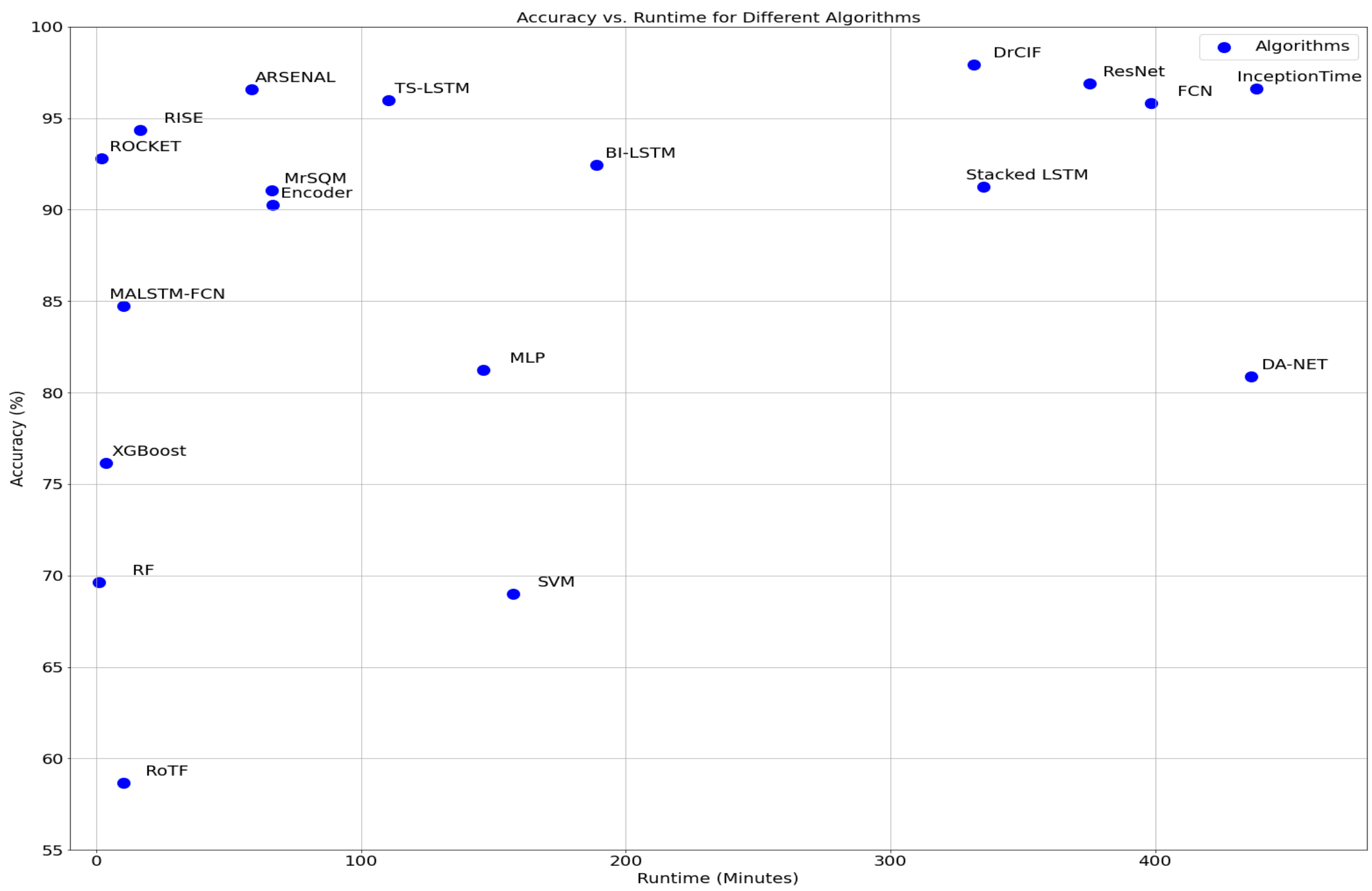

Figure 19: Average Runtime vs. Accuracy for 19 algorithms on 22 datasets. The horizontal axis is the average runtime of each algorithm in minutes and the vertical axis is the average accuracy of each algorithm in percentage

These results must be considered an indicator only for the scalability of tested algorithms as they have not been obtained based on a rigorous methodology but as additional results of the main experiment. Scalability is a very important factor to consider when choosing the best algorithm for a given problem. If the problem is associated with a large dataset or we anticipate the need to scale up the system in the future, we must consider the classifier's scalability. The upper-left part of Figure 19 showcases algorithms that exhibit commendable scalability while maintaining an acceptable level of accuracy. Algorithms like ARSENAL, ROCKET, and RISE fall into this category. Conversely, the upper-right part of the figure presents algorithms such as InceptionTime, FCN, and ResNet, where high accuracies come at the expense of extended runtimes. Some algorithms with exceptionally high runtimes were excluded from this figure. Notably, algorithms like HIVE-COTE V2.0 belong to this group and may encounter scalability issues to a degree that renders them impractical for certain applications.

Finally, in Figure 20, the scatter plot for the five top-performing algorithms (i.e., ARSENAL, DrCIF, TS-LSTM, ResNet, and Inception) runtime minutes vs accuracy percentage was plotted. In the upper part of the figure, each algorithm has 22 data points (marked with color-coded circles). In the lower part of the figure, the points with an accuracy of less than 90% and runtime higher than 1,000 minutes were removed to provide a zoomed-in comparison of the algorithms and their variabilities. The ellipses are drawn to mark the 95% confidence interval around the mean (marked with the star), assuming a bivariate normal distribution of points in the horizontal and vertical axis meaning 95% of the data points are in each respective ellipse. The scatter plot shows that ARSENAL and TS-LSTM have the minimum variability in both accuracy and runtime metrics among other algorithms, making them good candidates when performance stability is needed. In contrast, ResNet and InceptionTime algorithms both have high accuracy and runtime variabilities. Using them requires caution and they are less reliable in this regard.

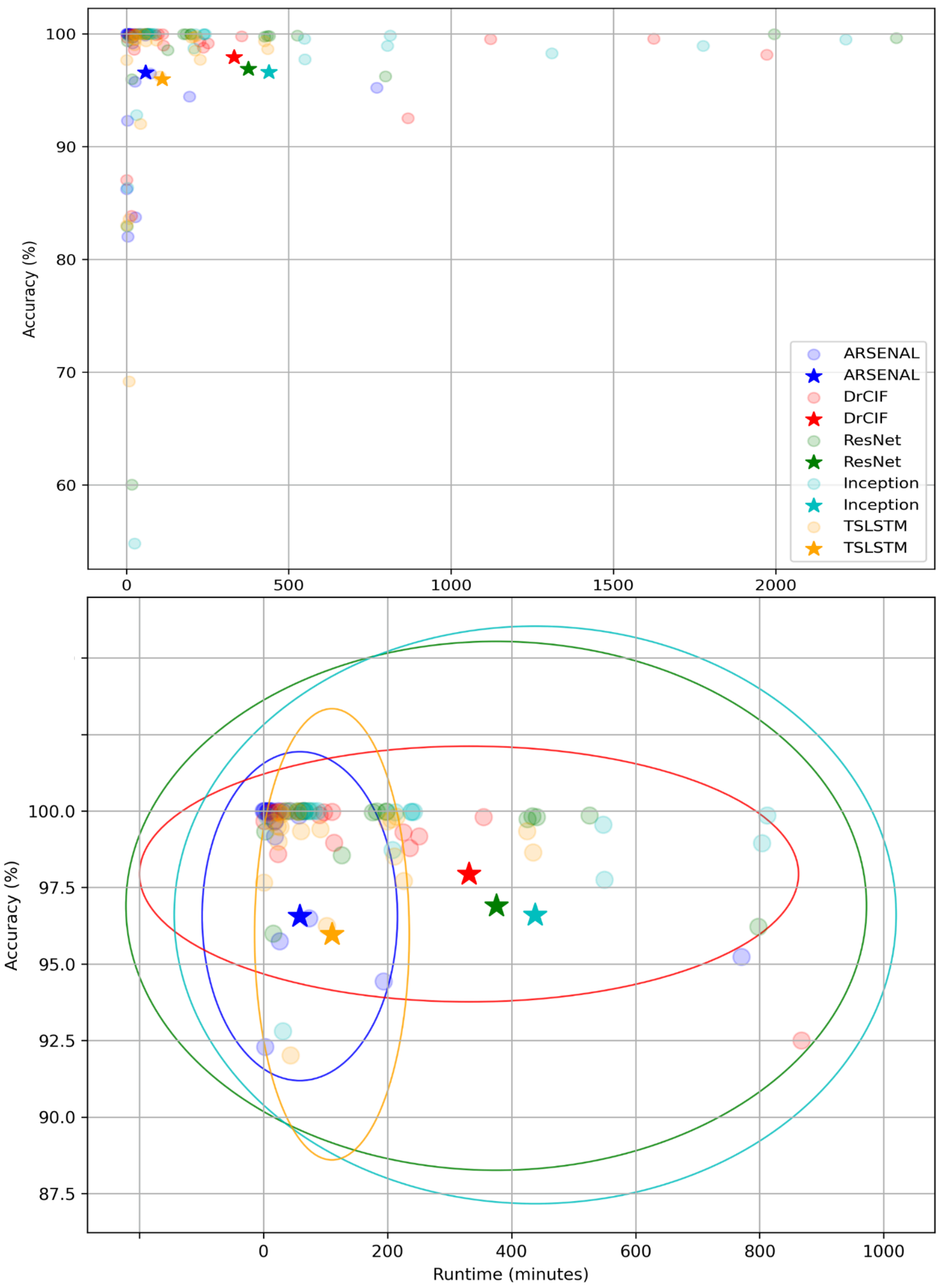

Figure 20: Scatterplot of all runtime vs. accuracies for the five top-performing algorithms (up). 95% confidence interval ellipses (down). The horizontal axis is the runtime of each algorithm in minutes and the vertical axis is the accuracy of each algorithm in percentage. Average accuracy and runtime are depicted by the star markers.

## 4.7. Practical Implications

Accuracy and runtime were discussed in detail in previous sections. However, when TSC algorithms are selected for manufacturing settings in practical applications, there are several other implications to consider besides high accuracy and low runtime. Data requirements, ease of implementation, required computational resources, and model interpretability are among those and are briefly discussed in the following.

First, before choosing any TSC algorithm for the use case, the *data* quality, quantity, and the needed effort to preprocess it should be considered as these factors can affect the model's accuracy regardless of the chosen algorithm. For example, DL models typically need larger datasets to perform well and avoid overfitting and in manufacturing environments, running the machines to collect more data can be very expensive and even infeasible in some cases. In this situation, conventional ML approaches that can work with less data might be more effective.

Second, the ease of *implementation* should be considered when we are choosing a TSC algorithm. Choosing an easy-to-deploy, open-sourced, and well-documented algorithm that requires minimal parameter tuning and technical expertise is favorable for practitioners and saves implementation time. Moreover, DL models need GPUs and considerably higher computational power to run efficiently. Thus, the needed computational *resources* should also be considered.

Finally, it is very important to choose a TSC algorithm providing interpretable predictions. This is a very important factor in manufacturing use cases because the prediction results are often used to conduct root cause analysis and fault diagnosis, to help with the continuous process improvement. For instance, DT-based classification algorithms, which may have lower accuracy, may produce more valuable predictions in sensitive cases compared to a highly accurate CNN model operating as a “black box” due to the clear understanding of how the predictions were made.

# 5. Conclusions, Future Work, and Limitations

Manufacturing industries are in dire need of AI and ML platforms to facilitate their transition towards Industry 4.0 and smart manufacturing systems, where a disconnect exists between the state-of-the-art ML algorithms in computer science literature and those utilized in manufacturing literature. Furthermore, practitioners within the manufacturing domain may lack the technical knowledge required to navigate the increasing number of algorithms, each with slight variations in structure and performance. This challenge is compounded by the multitude of parameters and hyperparameters that need tuning after selecting algorithms. In our efforts, we have undertaken exhaustive legwork, conducting experiments across various scenarios, and introducing algorithms that demonstrate strong out-of-the-box performance, particularly on manufacturing datasets. In doing so, we face yet another challenge in time-series analytics for smart manufacturing applications which is the scarcity of applicable public datasets, with only limited preprocessed manufacturing datasets accessible to both researchers and practitioners. Consequently, ML researchers in manufacturing either must rely on original data collected from machines, which can be challenging and not feasible in many cases, or start from scratch with data preprocessing tailored to their specific applications. In a recent effort, we provided a structured overview of the current state of time-series pattern recognition in manufacturing, emphasizing practical problem-solving approaches. Building upon this foundation, here in this study, we have developed a specialized ML framework for TSC in smart manufacturing systems, empowering manufacturers to address diverse challenges within the industry.

In this paper, we present the largest empirical study of TSC algorithms in the manufacturing domain to-date. The entire experiment required nearly two years of machine runtime, which we parallelized to ensure feasibility within our resource limitations. Based on our results, ResNet, DrCIF, InceptionTime, and ARSENAL emerged as the top-performing algorithms, boasting an average accuracy of over 96.6% across all 22 datasets. Notably, DrCIF and ARSENAL belong to the conventional ML algorithms category, highlighting that DL algorithms are not always the optimal choice and powerful alternatives exist. These findings underscore the robustness, efficiency, scalability, and effectiveness of convolutional kernels in capturing temporal features in time-series data collected from manufacturing systems for TSC tasks, as three out of the top four performing algorithms leverage these kernels for feature extraction. Additionally, LSTM, BiLSTM, and TS-LSTM algorithms deserve recognition for their effectiveness in capturing features within time-series data using recurrent RNN-based structures. It is important to note that these results were derived empirically on manufacturing time-series data for the defined scope and this study was not designed to draw any theoretical conclusions beyond the scope.

We also used runtime as a supplementary metric to help the decision-making process, shedding light on the trade-offs between accuracy and computational efficiency.

There are several topics and subtopics that we did not discuss in this work that can be considered in *future works*. These topics include but are not limited to topics such as addressing time-series data with variable length for TSC, addressing very long time-series collected from high-frequency systems, investigating the impact of different normalization techniques on the performance of TSC models in manufacturing, investigating recent generative AI, transformer-based, and LLM-based techniques and algorithms for TSC tasks in manufacturing, and more. These unexplored areas present opportunities for further investigation and advancement in this growing field.

There are several *limitations* that must be considered when interpreting the results of this paper. This paper was written under certain assumptions, timelines, and resource limitations, with a primary focus on providing a comprehensive TSC framework in smart manufacturing systems. While the complete removal of subjectivity and biases is impossible and arguably not desirable, we intend to maintain transparency by articulating the process and methodology used, enabling our audience to understand our biases, intent, understanding, and their influence on the content of this paper. In particular, the following limitations are worth noting:

In the methodology section, it was assumed that newer algorithms within the same classification categories would outperform their predecessors. While this assumption is intuitively correct, It is worth acknowledging that it may not hold true in all cases, as some specific algorithms developed for specific problems may exist that refute this assumption. However, studies like this involve making a number of decisions about how information is collected and analyzed, how experiments are conducted, etc. Often there is no one "correct" approach. Instead, we focused on being as transparent as possible in explaining all the steps to increase the clarity and reproducibility of our work. We adopted this assumption to help the downselection of representative algorithms within each category and to manage the scope of the experiment, as it was not feasible to run all the initial 92 algorithms on all datasets within a reasonable timeframe.

The recorded runtimes were calculated under the condition that five algorithms were run simultaneously on parallel CPU cores. It is important to acknowledge that this concurrent execution may impact the runtimes in certain cases. Results might have varied if we had the resources to run algorithms individually.

Although we tried to include as many TSC algorithms as possible, it is plausible that some algorithms (especially brand-new algorithms) were not included due to the timing of the search and the continuous and fast-paced nature of the research in this field. Nevertheless, the developed methodology can be applied to these new algorithms in the future to expand on the presented results. Moreover, although different evaluation metrics had been utilized in this study, there might be some limitations and biases in the used evaluation metrics. Finally, the results discussed in this paper are only applicable to the studied domain and problem and the findings are not generalizable to other domains.

# Acknowledgment

This material is based upon work supported by the National Science Foundation under Grant No. 2119654. Any opinions, findings, conclusions, or recommendations expressed in this material are those of the author(s) and do not necessarily reflect the views of the National Science Foundation.

The authors express their appreciation for the contribution and valuable discussions with Robert Gianinny in the early stages of the study that improved the quality of this paper. The authors express their appreciation to the Robotics and Computer-Integrated Manufacturing reviewers for their feedback and resulting improvements.